\PassOptionsToPackage{prologue,dvipsnames}{xcolor}
\documentclass[acmtog]{acmart}
\acmSubmissionID{2061}
\usepackage{booktabs} 
\usepackage[ruled]{algorithm2e} 

\SetAlFnt{\small}
\SetAlCapFnt{\small}
\SetAlCapNameFnt{\small}
\SetAlCapHSkip{0pt}

\acmJournal{TOG}

\usepackage[utf8]{inputenc} 
\usepackage[T1]{fontenc}    
\usepackage{hyperref}       
\usepackage{url}            
\usepackage{booktabs}       
\usepackage{amsfonts}       
\usepackage{nicefrac}       
\usepackage{microtype}      
\usepackage{xcolor}         
\usepackage{graphicx}
\usepackage{amsmath}
\usepackage{adjustbox}
\usepackage{xspace}
\usepackage{multicol}
\usepackage{multirow}
\usepackage[dvipsnames]{xcolor}
\usepackage{colortbl}
\usepackage{caption}
\usepackage{subcaption}
\usepackage{enumitem}
\usepackage{xcolor} 
\usepackage{fontawesome5}
\usepackage{tabularx}
\usepackage{array}
\usepackage{makecell} 
\usepackage{amsmath} 
\usepackage{fontawesome5} 
\usepackage{mdframed}
\begin{document}

\title{\name: A Unified Framework for Motion Synthesis and Comprehension}
\settopmatter{printacmref=false}

\newcommand{\figref}[1]{\mbox{Fig~\ref{#1}}}
\newcommand{\tabref}[1]{\mbox{Tab~\ref{#1}}}
\newcommand{\secref}[1]{\mbox{Sec~\ref{#1}}}
\newcommand{\appsecref}[1]{\mbox{Appendix Sec~\ref{#1}}}
\newcommand{\name}{VersatileMotion\xspace}
\newcommand{\vqname}{FlowVQ\xspace}
\newcommand{\abbr}{VerMo\xspace}

\author{Zeyu Ling}
\authornote{Both authors contributed equally to this research.}
\email{zeyuling@zju.edu.cn}
\affiliation{%
  \institution{Zhejiang University}
  \city{Hangzhou}
  \state{Zhejiang}
  \country{China}
}

\author{Bo Han}
\authornotemark[1]
\email{borishan.815@bytedance.com}
\affiliation{%
  \institution{ByteDance}
  \city{Hangzhou}
  \state{Zhejiang}
  \country{China}
}

\author{Shiyang Li}
\email{22221192@zju.edu.cn}
\affiliation{%
  \institution{Zhejiang University}
  \city{Hangzhou}
  \state{Zhejiang}
  \country{China}
}

\author{Jikang Cheng}
\email{chengjikang23@mails.ucas.ac.cn}
\affiliation{%
  \institution{University of the Chinese Academy}
  \city{Hangzhou}
  \state{Zhejiang}
  \country{China}
}

\author{Hongdeng Shen}
\email{shenhongdeng23@mails.ucas.ac.cn}
\affiliation{%
  \institution{University of the Chinese Academy}
  \city{Hangzhou}
  \state{Zhejiang}
  \country{China}
}

\author{Changqing Zou}
\authornotemark[2]
\authornote{Corresponding author}
\email{changqing.zou@zju.edu.cn}
\affiliation{%
  \institution{Zhejiang Lab}
  \city{Hangzhou}
  \state{Zhejiang Province}
  \country{China}
  }

\begin{teaserfigure}
    \centering
    \includegraphics[width=\textwidth]{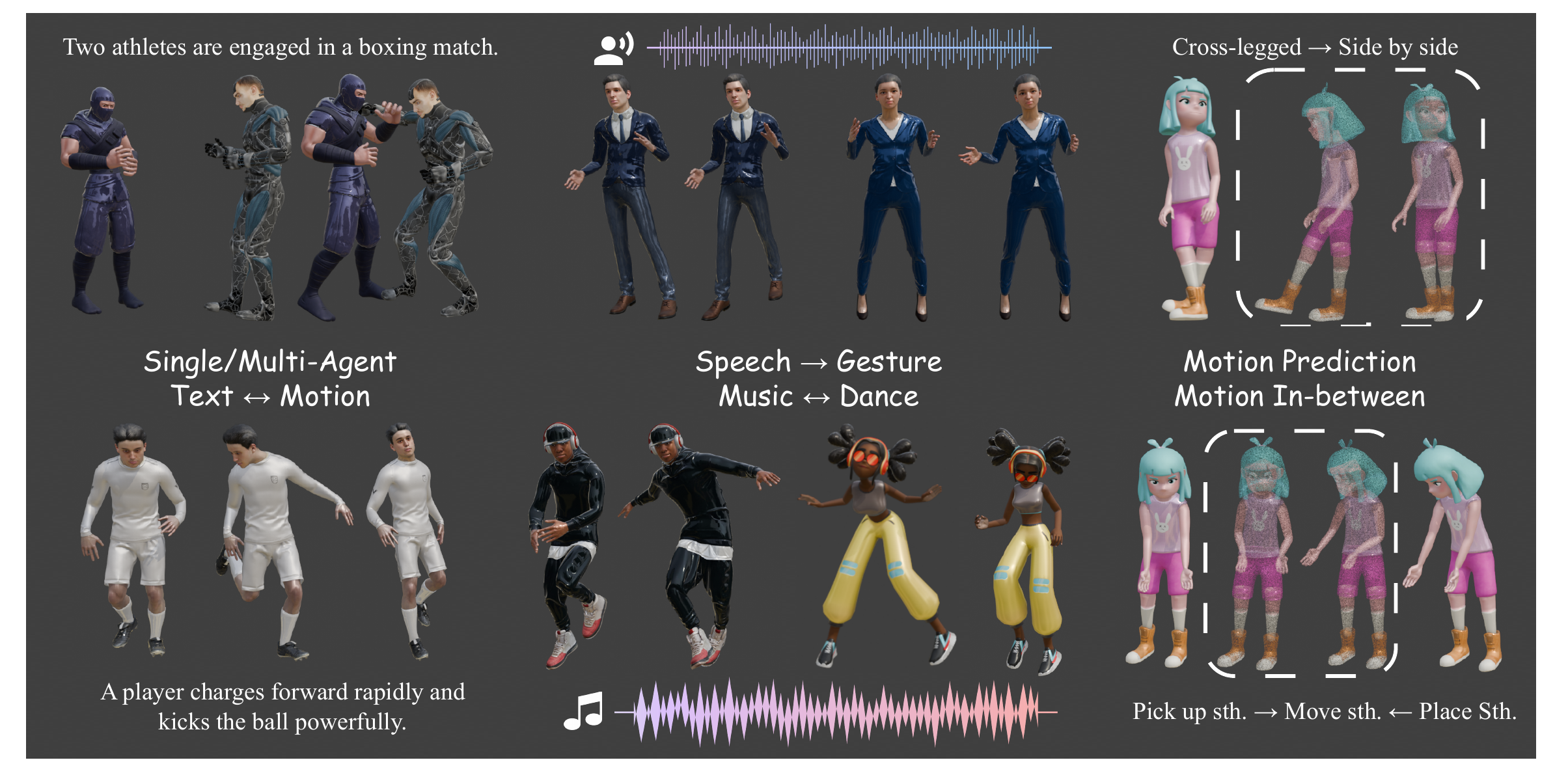}
    \caption{\name supports both single‑agent and multi‑agent motion synthesis and understanding, enabling seamless cross‑modal conversion among text, audio, and motion (both single and multi‑agent).}   
    \label{fig:main_fig} 
\end{teaserfigure}

\begin{abstract}

Large language models (LLMs) are, by design, inherently capable of multi-task learning: through a unified next-token prediction paradigm, they can naturally address a wide variety of downstream tasks. Prior work in the motion domain has demonstrated some generality by adapting LLMs via a Motion Tokenizer coupled with an autoregressive Transformer to generate and understand human motion. However, this generality remains limited in scope and yields only modest performance gains. We introduce \name, a unified multimodal motion LLM that combines a novel motion tokenizer, integrating VQ-VAE with flow matching, and an autoregressive transformer backbone to seamlessly support at least nine distinct motion-related tasks. \name is the first method to handle single-agent and multi-agent motions in a single framework and enable cross-modal conversion between motion, text, music, and speech, achieving state-of-the-art performance on seven of these tasks. Each sequence in MotionHub may include one or more of the following annotations: natural-language captions, music or audio clips, speech transcripts, and multi-agent interaction data. To facilitate evaluation, we define and release benchmark splits covering nine core tasks. Extensive experiments demonstrate the superior performance, versatility, and potential of \name as a foundational model for future understanding and generation of motion. 

\begin{figure*}[htbp]
    \centering
    \includegraphics[width=0.9\textwidth]{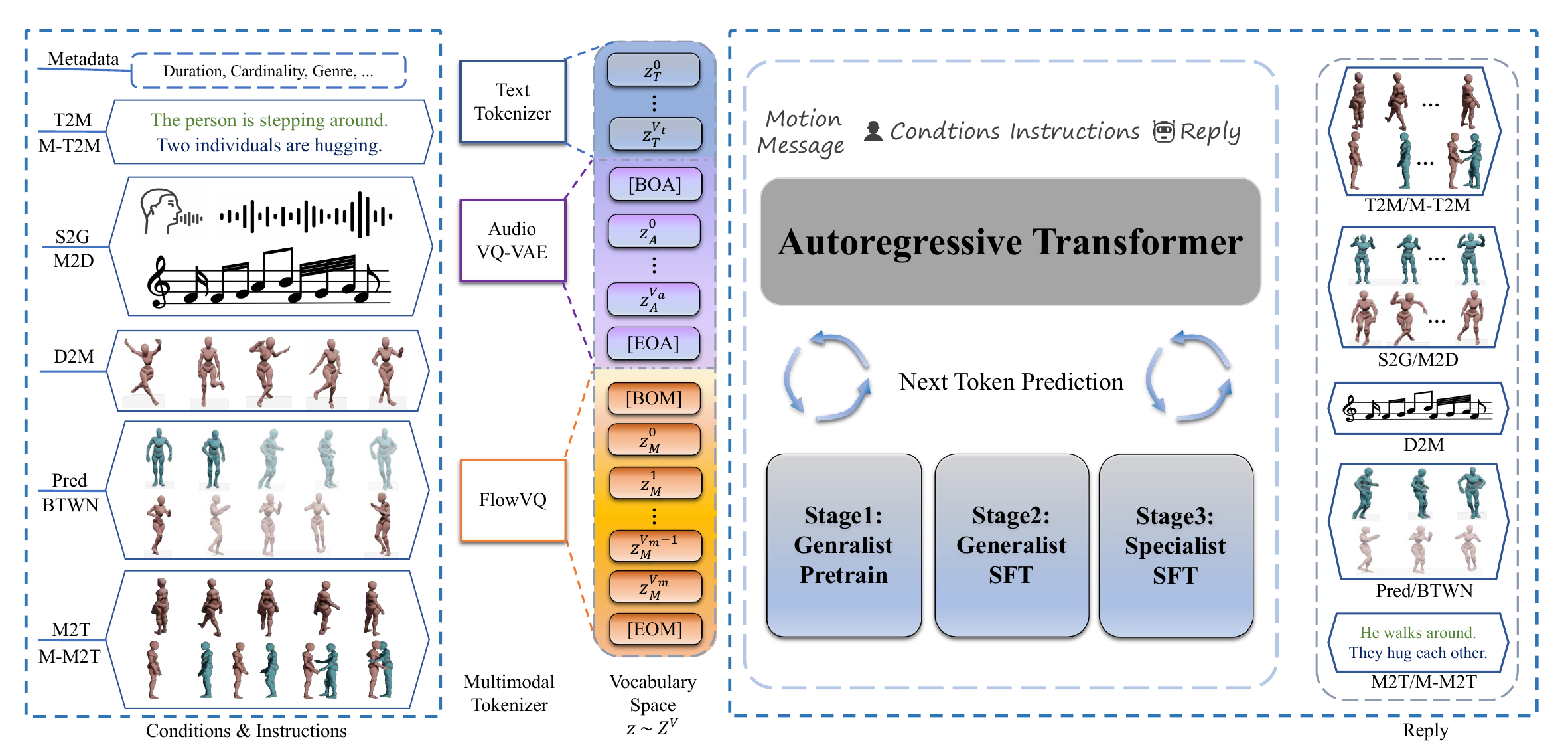}
    \caption{Overview of \name. 
    The training follows a three-stage "generalist to specialist" strategy. Each training sample consists of a condition (such as caption, audio, metadata, etc.), an instruction (briefly describing the task objective, e.g., generating a motion sequence based on a caption), and a reply (the model's output). Each sample is referred to as a motion message.}
    \label{fig:overview}
\end{figure*}

\end{abstract}

\maketitle

\begin{table}[htbp]
    \centering
    \caption{The unified multitasking ability of the proposed \name. \textit{T2M}: text-to-motion, \textit{M2D}: music-to-dance, \textit{S2G}: speech-to-gesture, \textit{M-T2M}: multi-agent text-to-motion, \textit{M2M}: including motion prediction and in-between, \textit{M2T}: motion-to-text, \textit{D2M}: dance-to-music, \textit{M-M2T}: multi-agent motion-to-text.}
    \adjustbox{width=0.95\linewidth}{
    \begin{tabular}{ccccccccc}
    \toprule
    \multirow{2}{*}{\centering Methods} & \multicolumn{5}{c}{Synthesis} & \multicolumn{3}{c}{Comprehension}\\
    \cmidrule(r){2-6} \cmidrule(r){7-9}
     & \textit{T2M} & \textit{M2D} & \textit{S2G} & \textit{M-T2M} & \textit{M2M} & \textit{M2T} & \textit{D2M} & \textit{M-M2T} \\
    \midrule
    MDM~\cite{mdm} & \checkmark & $\times$ & $\times$ & \checkmark & $\times$ & $\times$ & $\times$ & $\times$\\
    TM2T~\cite{tm2t} & \checkmark & $\times$ & $\times$ & $\times$ & $\times$ & \checkmark & $\times$ & $\times$ \\
    TM2D~\cite{tm2d} & \checkmark & \checkmark & $\times$  & $\times$ & $\times$ & $\times$ & $\times$ & $\times$\\
    UDE~\cite{ude} & \checkmark & \checkmark & $\times$ & $\times$ & $\times$ & $\times$ & $\times$ & $\times$ \\
    MCM~\cite{mcm} & \checkmark & \checkmark & \checkmark & $\times$ & $\times$ & $\times$ & $\times$ & $\times$ \\
    MotionGPT~\cite{motiongpt} & \checkmark & $\times$ & $\times$ & $\times$ & \checkmark & \checkmark & $\times$ & $\times$\\
    LMM~\cite{lmm} & \checkmark & \checkmark & $\times$ & $\times$ & \checkmark & $\times$ & $\times$ & $\times$\\
    UDE2~\cite{ude2} & \checkmark & \checkmark & \checkmark & $\times$ & $\times$ & $\times$ & $\times$ & $\times$\\
    UniMuMo~\cite{unimumo} & $\times$ & \checkmark & $\times$ & $\times$ & $\times$ & $\times$ & \checkmark & $\times$\\
    LoM~\cite{language_of_motion} & \checkmark & $\times$ & \checkmark & $\times$ & $\times$ & $\times$ & $\times$ & $\times$ \\
    M$^3$GPT~\cite{m3gpt} & \checkmark & \checkmark & $\times$ & $\times$ & \checkmark & \checkmark & \checkmark & $\times$ \\
    MotionAgent~\cite{motionllm} & \checkmark & $\times$ & $\times$ & \checkmark & $\times$ & \checkmark & $\times$ & $\times$ \\ \midrule 
    \name & \checkmark & \checkmark & \checkmark & \checkmark & \checkmark & \checkmark & \checkmark  & \checkmark \\
    \bottomrule
    \end{tabular}
    }
    \label{tab:tasks_comparison}
\end{table}

\begin{table*}
    \centering
    \caption{The constituent subsets of \textbf{MotionHub}. All values reflect post-processed data—after operations such as slicing, annotation, filtering, and resampling. Superscripts indicate caption annotation granularity: \textsuperscript{1}~Macro-level, \textsuperscript{2}~Meso-level, \textsuperscript{3}~Micro-level.}
    \adjustbox{width=0.9\textwidth}{
    \begin{tabular}{c|cccccc}
    \toprule
     \multirow{2}{*}{\centering Subset} & \multirow{2}{*}{\makecell{Duration \\ (Hours)}} & \multicolumn{2}{c}{Motion Clips (Instances)} & \multicolumn{2}{c}{Caption (Instances)} & \multirow{2}{*}{\makecell{Audio\\(Instances)}} \\
    \cmidrule(r){3-4} \cmidrule(r){5-6}
    & & Single & Multi & Single & Multi \\
    \midrule
    CNRS~\cite{AMASS_KIT-CNRS-EKUT-WEIZMANN,AMASS_KIT-CNRS-EKUT-WEIZMANN-2,AMASS_KIT-CNRS-EKUT-WEIZMANN-3} & 0.15 & 78 & - & 78\textsuperscript{[3]} & - & - \\  
    DanceDB~\cite{AMASS_DanceDB} & 6.74 & 2,012 & - & 2,012\textsuperscript{[3]} & - & - \\  
    HUMAN4D~\cite{AMASS_HUMAN4D} & 1.91 & 652 & - & 652\textsuperscript{[3]} & - & - \\  
    MOYO~\cite{moyo} & 2.73 & 936 & - & 936\textsuperscript{[3]} & - & - \\  
    SOMA~\cite{AMASS_SOMA} & 0.47 & 174 & - & 174\textsuperscript{[3]} & - & - \\  
    WEIZMANN~\cite{AMASS_KIT-CNRS-EKUT-WEIZMANN,AMASS_KIT-CNRS-EKUT-WEIZMANN-2,AMASS_KIT-CNRS-EKUT-WEIZMANN-3} & 7.85 & 3,017 & - & 3,017\textsuperscript{[3]} & - & - \\  
    PerMo~\cite{permo} & 8.56 & 6,610 & - & 6,610\textsuperscript{[2,3]} & - & - \\
    HumanAct12~\cite{humanact12,phspd} & 2.71 & 2,570 & - & 7,045\textsuperscript{[1,2,3]} & - & - \\
    UESTC~\cite{uestc} & 109.32 & 36,599 & - & 36,599\textsuperscript{[1]} & - & - \\
    Human3.6M~\cite{human36m} & 3.02 & 939 & - & 939\textsuperscript{[1,2]} & - & - \\
    NTU-RGBD-120~\cite{nturgbd} & 75.86 & 114,480 & - & 114,480\textsuperscript{[1]} & - & - \\
    Fit3D~\cite{fit3d} & 3.13 & 944 & - & 2,795\textsuperscript{[2,3]} & - & - \\
    HumanSC3D~\cite{humansc3d} & 1.19 & 688 & - & 2,086\textsuperscript{[2,3]} & - & - \\
    Trumans~\cite{trumans} & 6.88 & 3,624 & - & 3,624\textsuperscript{[1,2,3]} & - & - \\
    3DPW~\cite{3dpw} & 0.48 & 60 & - & 60\textsuperscript{[1,2]} & - & - \\
    MotionX~\cite{motionx,motionx++} & 165.56 & 85,758 & - & 105,952\textsuperscript{[1,2,3]} & - & 3,392 \\
    Chi3D~\cite{chi3d} & 0.71 & 746 & 373 & 1,409\textsuperscript{[2,3]} & 211\textsuperscript{[2,3]} & - \\
    HI4D~\cite{hi4d} & 0.22 & 200 & 100 & 984\textsuperscript{[1,2,3]} & 522\textsuperscript{[1,2,3]} & - \\
    InterHuman~\cite{intergen} & 20.77 & 15,544 & 7,772 & 24,964\textsuperscript{[2,3]} & 23,310\textsuperscript{[2,3]} & - \\
    InterX~\cite{interx} & 37.49 & 22,776 & 11,388 & 68,382\textsuperscript{[2,3]} & 34,194\textsuperscript{[2,3]} & - \\
    AIST++~\cite{fact} & 5.20 & 1,408 & - & 1,408\textsuperscript{[3]} & - & 1,408 \\
    FineDance~\cite{finedance} & 14.12 & 4,237 & - & 4,237\textsuperscript{[3]} & - & 4,237 \\
    BEATv2~\cite{beatv1,emage} & 61.34 & 23,251 & - & - & - & 23,251 \\
    TED~\cite{tedexpressive,tedgesture} & 60.05 & 31,544 & - & - & - & 31,544 \\
    \midrule
    MotionHub & \textbf{596.48} & \textbf{358,847} & \textbf{19,633} & \textbf{388,443\textsuperscript{[1,2,3]}} & \textbf{58,237\textsuperscript{[1,2,3]}} & \textbf{63,832} \\
    \bottomrule
    \end{tabular}
    }
    \label{tab:motionhub_comp}
\end{table*}

\section{Introduction}

Motion synthesis and comprehension are fundamental problems in computer vision and graphics, which underpin applications from virtual reality and animation to human–computer interaction and biomechanics. Deep learning methods—especially VAEs, GANs, and diffusion models—have driven remarkable advances in generating realistic human motion, while sequence-to-sequence models and retrieval frameworks have improved the ability to interpret motion signals. However, most prior works treat synthesis and comprehension as separate paradigms, and few systems can flexibly handle multiple modalities or multi-agent scenarios.

Recent breakthroughs in LLMs have demonstrated their power as universal multitask learners in NLP~\cite{t5,flant5,gpt3}. Inspired by this, several studies convert motion into discrete token sequences via VQ–VAE and apply Transformers to text-conditioned generation, captioning, and interpolation~\cite{motiongpt,t2mgpt,m3gpt}. Frameworks such as MotionGPT~\cite{motiongpt,motiongpt2} and MotionAgent~\cite{motionagent} unify single-agent tasks, and \(M^3\)GPT~\cite{m3gpt} extends this to bidirectional music–dance translation. However, two key limitations remain unexplored:

\begin{itemize}[leftmargin=5mm, label=\normalfont\textbullet, itemsep=0.5ex, topsep=0.5ex, parsep=0ex]
    \item \textbf{Limited Versatility.} Existing approaches have primarily focused on the understanding and generation of single‑agent motion, with only limited exploration of motion‑related audio modalities (e.g., music, voice) for both comprehension and synthesis.

    \item \textbf{Quality Bottlenecks.} Previous approaches---such as enhanced quantizers (e.g., FSQ~\cite{fsq,scamo}) and residual-token hierarchies paired with multi-stage Transformers~\cite{rvq,m3gpt,t2mhifigpt,momask}---partially alleviate VQ--VAE's quantization noise, but achieve only modest gains in generation fidelity and fall short in capturing intricate, fine-grained motion details.
\end{itemize}

To overcome these limitations, we propose \name, a unified Multimodal Motion LLM built on three core principles:

First, at the heart of \name lies \vqname, a novel Motion Tokenizer that marries discrete VQ–VAE quantization with a Flow Matching Transformer decoder: by first mapping continuous motion sequences into a compact codebook of tokens and then learning a time‐conditioned vector field to iteratively denoise and refine the reconstructed trajectories, \vqname dramatically reduces quantization artifacts and captures high‐frequency motion nuances that elude standard VQ–VAE decoders.

Second, we introduce MotionHub, the largest standardized motion corpus to date: over 350 K SMPL-H sequences drawn from heterogeneous mocap sources are unified via automated retargeting and monocular recovery, annotated with high‐quality manual captions or automated pipelines, aligned music and speech tracks, and multi‐agent interaction metadata.  MotionHub underpins nine benchmark tasks—ranging from single and multi-agent text $\leftrightarrow$ motion generation to music $\leftrightarrow$ dance and speech $\rightarrow$ gesture translation—and provides the scale and consistency necessary for robust pretraining.

Third, we introduce a three-stage generalist-to-specialist training paradigm. A \textbf{generalist} model is first obtained via pretraining and instruction tuning on a diverse mix of motion-centric tasks spanning single-/multi-agent settings and multiple modalities (text, audio). This foundation model already surpasses many task-specific baselines.

We then instruction-tune it on individual domains to yield \textbf{specialist} variants, which achieve new state-of-the-art results on 7 out of 9 motion benchmarks. To our knowledge, \name is the first unified framework supporting both single- and multi-agent motion, enabling seamless cross-modal generation across motion, text, music, and speech.

Overall, our contributions are threefold: first, we introduce \vqname, a novel Motion Tokenizer that integrates discrete VQ–VAE quantization with a Flow Matching Transformer decoder to produce high-fidelity motion tokens; second, we develop a generalist $\rightarrow$ specialist training paradigm in which a single generalist Motion LLM—trained across diverse motion, text, and audio tasks—already surpasses many task-specific baselines out of the box, and can then be fine-tuned into specialist models that achieve new state-of-the-art performance on each of nine benchmark tasks; and third, we assemble MotionHub, the largest unified multimodal motion dataset to date with over 350 K SMPL-H sequences for nine standardized tasks. Together, these advances enable \name to deliver both unprecedented versatility and generation fidelity, setting new SOTA on seven of nine evaluated tasks.

\section{Related Work}

\label{sec:related_work}
\subsection{Motion synthesis and comprehension}
Motion-related tasks can be broadly categorized into motion synthesis (e.g., text-to-motion, music-to-dance, speech-to-gesture) and motion comprehension (e.g., motion-to-text, dance-to-music). Below, we outline the key characteristics and recent advancements in each area.

Text-to-motion aims to generate motions that align with the semantic descriptions. 
Recent studies have primarily employed diffusion models~\cite{mdm,mld,mcm} 
and autoregressive Transformer models~\cite{humantomato,t2mgpt,momask} to achieve.
Music-to-dance focuses on generating dance motions that are consistent with the rhythm of the music. 
Unlike semantic signals, musical signals do not provide explicit descriptions, resulting in greater freedom~\cite{zhu2023human}. 
From a modeling perspective, recent approaches have converged with text-to-motion methodologies, utilizing diffusion~\cite{edge,finedance,alexanderson2023listen} and autoregressive models~\cite{bailando,fact,danceformer,amd}. However, they adopt different logic for integrating conditional signals.
Speech-to-gesture emphasizes generating facial and hand motions that correspond to speech. Compared to the aforementioned motion generation tasks, this task requires more refined modeling of hand and facial movements, often separating these parts from the torso for more precise modeling~\cite{talkshow,emage,beat}. Although these works have achieved promising results, they are often limited to a single task.

In dance-to-music tasks, music is typically represented as discrete tokens. This can be achieved through the inherently discrete notation of symbolic music or by discretizing waveform music. Foley~\cite{foley} focuses on generating symbolic music that aligns with performance motions, while Dance2MIDI~\cite{dance2midi} emphasizes dance motion, generating rhythm first and then conditionally generating notes for other instruments. Taking video as input, CMT~\cite{cmt} establishes three types of relationships between video and music, whereas D2M-GAN~\cite{d2m-gan} introduces a GAN-based approach for generating waveform music tokens. Similarly, this approach translates the generation task into a sequence-to-sequence translation problem in motion-to-text tasks~\cite{tm2t,goutsu2021linguistic,wu2023large,plappert2018learning}.

\subsection{Multimodal LLMs}
LLMs such as GPT-3~\cite{gpt3}, GPT-4~\cite{achiam2023gpt}, PaLM~\cite{palm}, and LLaMA~\cite{llama}, developed through extensive training on massive text datasets. By scaling both the dataset size and model architecture, LLMs exhibit remarkable emergent abilities, including instruction following~\cite{peng2023instruction}, In-Context Learning (ICL)~\cite{gpt3}, and Chain-of-Thought (CoT) reasoning~\cite{cot}.
Multimodal LLMs (MLLMs) extend these capabilities by integrating and processing information from multiple modalities. 
Since the release of GPT-4o~\cite{achiam2023gpt}, research on MLLMs has intensified, focusing initially on generating text from image~\cite{openflamingo}, video~\cite{video-llama}, and audio inputs~\cite{pengi}. Subsequent studies have broadened the scope to enhance support for diverse input and output modalities~\cite{llamaadapter,moon2023anymal}. 
Notably, NExT-GPT~\cite{wu2023next} utilizes multimodal adaptors and diffusion decoders, enabling it to perform tasks involving arbitrary combinations of text, images, videos, and audio.

\subsection{Motion LLMs}
Recently, several studies have attempted to apply MLLMs to the domain of human motion, aiming to address various motion-centric generative tasks. T2MGPT ~\cite{t2mhifigpt} is the first to integrate  T5~\cite{t5} into the motion domain, while MotionGPT~\cite{motiongpt} further enables mutual transformation between motion and text. LMM~\cite{lmm}, built upon the FineMoGen~\cite{finemogen}, introduces an MoE architecture for motion generation. Similarly, Language of Motion~\cite{language_of_motion}, M\(^3\)GPT~\cite{m3gpt} and UDE-2~\cite{ude2} also adopt T5 as its language model backbone and pairs it with a conditional modality tokenizer to establish a motion-focused LLM. UniMuMo~\cite{unimumo} primarily concentrates on dance movements, additionally supporting conversion between music and text. While these methods are capable of handling various motion-related generative tasks, they remain at the million level. 
The concurrent work, MotionAgent~\cite{motionllm}, achieves a billion-level scale, but it struggles with fine-grained motion synthesis, particularly in generating detailed hand motions.

\begin{figure}
    \centering
    \includegraphics[width=0.95\linewidth]{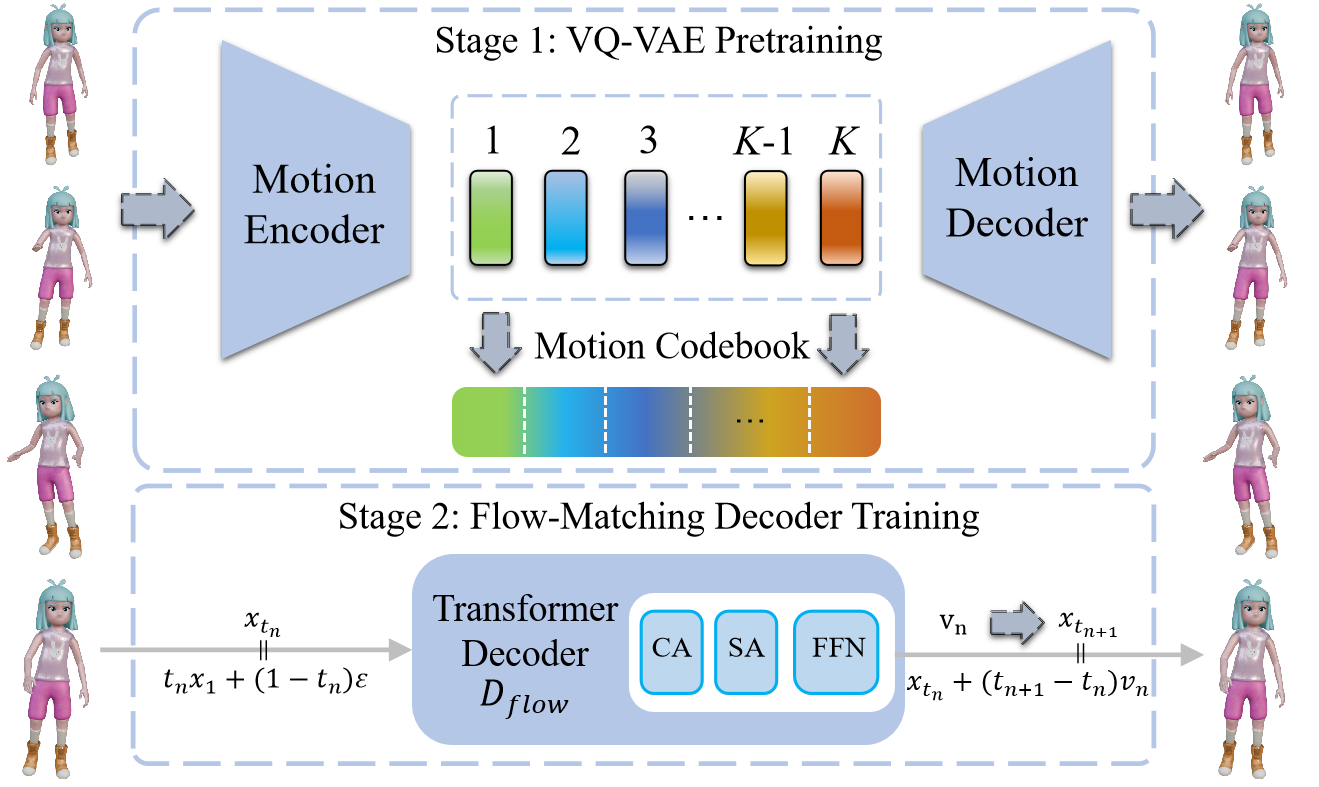}
    \caption{The schematic of \vqname.}
    \label{fig:flowvq}
\end{figure}
\section{MotionHub and MotionBench}

\subsection{Motivation for Curating MotionHub}
Training large‐scale generative models critically depends on vast, high‐quality data, yet motion benchmarks lag far behind image and video in both scale and consistency.  Existing datasets—even when covering similar tasks—differ in representation (e.g., HM3D ~\cite{t2m}, HumanTOMATO ~\cite{humantomato}, InterHuman ~\cite{intergen}), skeleton topology (e.g., COCO ~\cite{coco}, SMPL~\cite{smpl}, Human3.6M~\cite{human36m}), and coordinate conventions, impeding unified model training and evaluation.  Moreover, dataset sizes vary dramatically across tasks, preventing models from learning generalizable motion priors.  \textbf{MotionHub} remedies these gaps by aggregating over 400K sequences into a single SMPL-H format, enabling true large‐scale pretraining and standardized multi‐task benchmarks.

\definecolor{headerbg}{HTML}{D9EAF7}   
\definecolor{rowbg}{HTML}{F5F5F5}      

\begin{table*}[t]
    \centering
    \caption{Examples of different tasks and their corresponding motion message formats.}
    \vspace{0.5em}
    \adjustbox{width=0.9\linewidth}{
    \rowcolors{2}{rowbg}{white}
    \begin{tabular}{lcccc}
    \toprule
    \rowcolor{headerbg}
    \textbf{Task} & \textbf{Instruction} & \textbf{Required Condition(s)} & \textbf{Optional Condition(s)} & \textbf{Reply} \\
    \midrule
    \texttt{T2M}      & Generate a motion sequence based on the user's caption.               & Caption                      & Duration               & Motion \\
    \texttt{N2TM}     & Randomly synthesize a motion sequence and its caption.               & --                           & Duration               & Motion + Caption \\
    \texttt{M-T2M}    & Generate a motion sequence with a specified number of participants.  & Caption + Headcount          & Duration               & Motion \\
    \texttt{M2T}      & Describe the given motion.                                            & Motion                       & --                     & Caption \\
    \texttt{M-M2T}    & What are the persons doing?                                           & Motion + Headcount           & --                     & Caption \\
    \texttt{M-N2TM}   & Randomly generate a motion sequence with a specified number of participants. & Headcount              & Duration               & Caption + Motion \\
    \texttt{S2G}      & Add body movements to this speech.                                   & Speech                       & Caption                & Motion \\
    \texttt{G2S}      & Infer the speech audio based on the gestures.                        & Motion                       & --                     & Audio \\
    \texttt{N2SG}     & Randomly generate speech and corresponding body movements.           & --                           & Duration + Caption     & Audio + Motion \\
    \texttt{M2D}      & Dance to the given music.                                             & Audio                        & Caption + Genre        & Motion \\
    \texttt{D2M}      & Add background music to this dance.                                  & Motion                       & Genre                  & Audio \\
    \texttt{N2DM}     & Compose a piece of music and pair it with a dance.                   & --                           & Duration + Genre       & Audio + Motion \\
    \texttt{Pred}     & Predict future frames from past motion.                              & Past Motion                  & Duration + Caption     & Motion \\
    \texttt{BTWN}     & Interpolate missing frames between two motion segments.              & Past + Future Motion         & Duration + Caption     & Motion \\
    \bottomrule
    \end{tabular}
    }
    \label{tab:motion_message}
\end{table*}

\subsection{MotionHub Construction Pipeline}

\paragraph{$\triangleright$\hspace{0.4em}\textit{Unified Representation and Standardization}}
All raw motion is retargeted into SMPL-H and InterHuman formats via automated skeleton mapping and monocular capture.  Clips are then truncated to 12 seconds, resampled to 30 fps, and aligned in a right-handed coordinate system (y-plane at global minimum joint height; xz-plane origin at the first agent’s initial pose and facing).  This ensures consistent topology, timing, and spatial conventions.

\paragraph{$\triangleright$\hspace{0.4em}\textit{Multi‑Granularity Caption Annotation}}
Captions in MotionHub are organized into three levels of granularity—macro, meso, and micro—matched to clip complexity and existing resources.  Macro‑level labels (single words or brief phrases) are generated by converting simple action tags in NTU-RGBD and UESTC, and by extracting per‑agent descriptions from the dual‑agent captions in InterX and InterHuman.  Meso‑level captions (one to three sentences) are adopted verbatim from prior annotations in MotionX, PerMo, and HumanAct12 (single‑agent) as well as InterX and InterHuman (multi‑agent).  All remaining clips—particularly those with intricate gestures or interaction dynamics—were annotated by trained labelers at micro granularity, providing fine‑grained narratives of pose transitions, timing cues, and inter‑agent relations.  Highly complex datasets (TED, BEAT2) are reserved for future annotation.  Each subset’s caption levels are indicated in ~\tabref{tab:motionhub_comp}.

\paragraph{$\triangleright$\hspace{0.4em}\textit{Rigorous Quality Control}}
We apply three validation layers: (i) filter low-fidelity captures via joint-velocity and occlusion checks, with manual review of outliers; (ii) verify cross-modal alignment between text–motion and audio–motion via semantic and temporal consistency checks; and (iii) validate all captions with GPT-4o for grammar and logic, followed by human correction.

In total, MotionHub spans 25 subsets over 596.48 hours, containing 358,847 single-agent motions, 19,633 multi-agent motions, 63,832 audio clips, 388,443 single-agent captions, and 58,237 multi-agent captions.

\subsection{MotionBench: Benchmark Design and Development}
We define nine benchmarks on MotionHub for single and multi-agent synthesis and comprehension. ~\tabref{tab:motionhub_comp} maps subsets to tasks.  Each benchmark adopts established metrics and extractors, with outdated methods replaced by advanced alternatives.

For text–motion tasks, we follow the HumanTOMATO framework~\cite{humantomato} to train a contrastive text–motion retrieval (TMR) model, computing motion quality and text–motion alignment.

Other benchmarks—music \(\leftrightarrow\) dance, motion interpolation, motion prediction—use modality-appropriate metrics.  Detailed definitions, splits, and implementations are in ~\appsecref{sec:add_eval_metrics}.

\begin{table*}
    \centering
    \caption{Quantitative evaluation results of T2M and M-T2M on MotionHub. The upper half of the table corresponds to T2M, while the lower half corresponds to M-T2M. The symbol $\rightarrow$ indicates that the results are more favorable when they are closer to real motions. The number 256 means the R-precision is calculated within 256 samples per batch.} 
    \adjustbox{width=0.9\linewidth}{
    \begin{tabular}{lcccccc}
        \toprule[2pt]
        \multirow{2}{*}{\centering Methods} & \multicolumn{3}{c}{$\text{R-Precision}^{256}$ $\uparrow$} & \multirow{2}{*}{\centering FID $\downarrow$} & \multirow{2}{*}{\centering MM Dist $\downarrow$} & \multirow{2}{*}{\centering Diversity $\rightarrow$} \\
        \cmidrule(r){2-4}
        & Top 1 & Top 2 & Top 3 & & & \\
        \midrule[1pt]
        Real & 0.5247 & 0.6979 & 0.7716 & - & 0.9834 & 17.4524 \\
        TM2T~\cite{tm2t} & 0.2300 & 0.3480 & 0.4113 & 211.0022 & 1.1867 & 17.0454 \\
        MDM~\cite{mdm} & 0.2717 & 0.4134 & 0.4963 & 96.7136 & 1.1587 & \textbf{17.4689} \\
        MotionDiffuse~\cite{motiondiffuse}  & 0.2909 & 0.4296 & 0.5023 & 142.5879 & 1.1196 & 17.0975 \\
        MotionGPT~\cite{motiongpt} & 0.2972 & 0.4356 & 0.5150 & 105.8154 & 1.1285 & \underline{17.4310} \\
        MoMask~\cite{momask} & 0.2992 & 0.4429 & 0.5206 & 76.0154 & 1.1042 & 17.3906 \\
        \abbr-1B-Gen & 0.2918 & 0.4218 & 0.4889 & \underline{63.4194} & 1.1337 & 17.4017 \\
        \abbr-1B-Spe & \underline{0.3162} & \underline{0.4555} & \underline{0.5304} & 64.7033 & \underline{1.0914} & 17.2965 \\
        \quad + \vqname & \textbf{0.3224} & \textbf{0.4674} & \textbf{0.5441} & \textbf{62.8245} & \textbf{1.0843} & 17.2179 \\
        \midrule[1.5pt]
        Real & 0.6582 & 0.7814 & 0.8311 & - & 0.9599 & 23.3325 \\
        MDM~\cite{mdm} & 0.0636 & 0.1108 & 0.1422 & 743.4390 & 1.3502 & 22.8706 \\
        ComMDM~\cite{priormdm} & 0.0803 & 0.1349 & 0.1758 & 420.6489 & 1.3334 & 23.2476 \\
        InterGen~\cite{intergen} & 0.1827 & 0.2859 & 0.3620 & 296.8057 & 1.2573 & 23.2332 \\
        \abbr-1B-Gen & 0.3862 & 0.4636 & 0.5051 & 147.1982 & 1.1571 & \textbf{23.3433} \\
        \abbr-1B-Spe & \underline{0.5009} & \underline{0.5893} & \underline{0.6353} & \underline{138.2251} & \underline{1.0939} & 23.2228 \\
        \quad + \vqname & \textbf{0.5188} & \textbf{0.6051} & \textbf{0.6413} & \textbf{135.5801} & \textbf{1.0896} & \underline{23.2841} \\
        \bottomrule[2pt]
    \end{tabular}
    }
    \label{tab:eval_t2m}
\end{table*}

\section{\name Framework}
\subsection{Preliminary: Multimodal Tokenization \& Autoregressive Transformer}

This work builds upon and extends the standard multimodal tokenization + autoregressive (AR) Transformer paradigm.  Inputs across text, audio, and motion are first discretized into a shared vocabulary \(Z\).  Formally, let  
\[
X=\{X_{\mathcal{T}},X_{\mathcal{A}},X_{\mathcal{M}}\},\quad Z=\{Z_{\mathcal{T}},Z_{\mathcal{A}},Z_{\mathcal{M}}\},
\]  
where \(\mathcal{T},\mathcal{A},\mathcal{M}\) denote text, audio, and motion.  Text is mapped via a subword tokenizer \(T:X_{\mathcal{T}}\!\to\!Z_{\mathcal{T}}\), while for \(m\in\{\mathcal{A},\mathcal{M}\}\) a VQ–VAE \((E_m,Q,D_m)\) learns  
\[
x\;\xrightarrow{E_m}\;y\;\xrightarrow{Q}\;z\;\xrightarrow{D_m}\;\hat x,
\quad x\in X_m,\;z\in Z_m,
\]  
by minimizing the following object:
\[\mathbb{E}_{x\sim p_{X_m}}\|x-D_m(Q(E_m(x)))\|^2+\beta\|E_m(x)-\mathrm{sg}[Q(E_m(x))]\|^2.\]

An AR Transformer \(P_\theta\) then fits the discrete sequences \(z=(z_1,\dots,z_T)\in Z^T\) by maximizing next-token likelihood:  
\[
\mathcal{L}_{\mathrm{train}}
=-\sum_{t=1}^T\log P_\theta(z_t\mid z_{<t}).
\]  
At inference, generation proceeds by sampling  
\(\;z_t\sim P_\theta(\cdot\mid z_{<t}),\)  
producing \(\hat z=(\hat z_1,\dots,\hat z_{T'})\), which is detokenized back into the original modality. 

\subsection{FlowVQ: Enhancing VQ–VAE with a Flow Matching Decoder}

As demonstrated in ~\figref{fig:flowvq}, \vqname\ augments a standard VQ–VAE — composed of an encoder, a discrete codebook, and a reconstruction decoder — with a second, flow-based decoder built on Transformer blocks. This design retains the simplicity of VQ–VAE’s quantization while leveraging flow matching to recover fine motion details lost during discretization. The training of \vqname\ comprises two stages, each optimizing a distinct objective:

\paragraph{Stage 1: VQ–VAE Pretraining}  

In the first stage, we train a standard vector-quantized autoencoder on motion data. An encoder \(E\) maps each motion sequence \(x\) to a continuous latent vector, which is then quantized by a learned codebook \(Q(\cdot)\) of discrete embeddings:
\begin{equation}
z = Q\bigl(E(x)\bigr),
\end{equation}
where \(z\) is the resulting discrete token sequence. A decoder \(D\) (the Motion Decoder) then reconstructs the motion from \(z\). We optimize a reconstruction loss:
\begin{equation}
\mathcal{L}_{\mathrm{rec}} = \mathbb{E}_{x\sim\mathcal{D}}\bigl\lVert x - D(z)\bigr\rVert^2,
\end{equation}
We adopt the FSQ quantizer for \(Q\), and therefore \(\mathcal{L}_{\mathrm{rec}}\) does not include any additional commitment loss term.  
So that \(D(z)\) closely matches the original motion \(x\). After training, the encoder \(E\) and codebook \(Q\) define a discrete tokenizer: each motion is represented by the sequence \(z\) instead of high-dimensional real-valued vectors. This yields a faithful baseline reconstruction but can be \emph{coarse}, since VQ quantization may lose subtle motion nuances or introduce frame-wise jitter.

\paragraph{Stage 2: Flow-Matching Decoder Training}  
For reconstruction refinement, we freeze the trained encoder \(E\) and codebook \(Q\) from Stage~1 and train a Transformer-based decoder \(D_{\mathrm{flow}}\) (parameterized by \(\theta\)) with a flow-matching objective. The module \(D_{\mathrm{flow}}\) contains cross-attention (CA), self-attention (SA), and feed-forward (FFN) blocks as shown in ~\figref{fig:flowvq}.

Let \(z = Q(E(x))\) denote the discrete code sequence. We define a discrete set of time steps \(\{t_n\}_{n=0}^N\) with \(t_0 = 0\) and \(t_N = 1\). For each \(t_n\), we sample Gaussian noise \(\epsilon \sim \mathcal{N}(0,I)\) and form a noisy motion:
\begin{equation}
x_{t_n} = \sqrt{\alpha_{t_n}}\,x + \sqrt{1 - \alpha_{t_n}}\,\epsilon,
\end{equation}
where \(\alpha_{t_n}\) is a predefined noise schedule. The flow-matching decoder then predicts the velocity field \(v_n\) at each time step \(t_n\), conditioned on \(x_{t_n}\), \(z\), and \(t_n\):
\begin{equation}
v_n = D_{\mathrm{flow}}(x_{t_n}, z, t_n).
\end{equation}
We minimize the loss:
\begin{equation}
\mathcal{L}_{\mathrm{flow}} = \mathbb{E}_{x,\epsilon,n}\bigl\lVert v_n - (x - x_{t_n})\bigr\rVert^2,
\end{equation}
which encourages the model to predict the residual between the clean motion \(x\) and the noisy input \(x_{t_n}\), effectively learning the velocity field that transports \(x_{t_n}\) back to \(x\).

In summary, Stage~1 produces a discrete token representation \(z = Q(E(x))\) via VQ–VAE, and Stage~2 employs a flow-matching Transformer decoder \(D_{\mathrm{flow}}\) that learns to predict velocity fields to denoise corrupted motion trajectories conditioned on \(z\), thereby reconstructing high-fidelity motion. This hybrid approach combines the efficiency of quantized codes with the smoothness of a diffusion-like refinement.

\begin{table*}
\centering
\caption{The upper part of the table presents the results of the MotionHub M2T benchmark, while the lower part displays the results of the M-M2T benchmark.}
\adjustbox{width=0.9\linewidth}{
    \begin{tabular}{lccccccccc}
    \toprule[2pt]
    \multirow{2}{*}{\centering Methods} &
    \multicolumn{3}{c}{$\text{R-Precision}^{256}\uparrow$} &
    \multirow{2}{*}{\centering MMDist$\downarrow$} &
    \multicolumn{2}{c}{BLEU$\uparrow$} &
    \multirow{2}{*}{\centering ROUGE-L$\uparrow$} &
    \multirow{2}{*}{\centering CIDEr-D$\uparrow$} &
    \multirow{2}{*}{\centering BERT Score$\uparrow$} \\
    \cmidrule(r){2-4} \cmidrule(r){6-7}
      & Top 1 & Top 2 & Top 3 &  & @1 & @4 &  &  & \\
    \midrule[1pt]
    Real & 0.5109 & 0.6885 & 0.7650 & 0.9834 & - & - & - & - & - \\
    TM2T~\cite{tm2t} & 0.2701 & 0.3951 & 0.4650 & 1.1448 & 37.56 & 19.72 & 41.89 & 2.3147 & 89.66 \\
    MotionGPT~\cite{motiongpt} & 0.3782 & 0.5218 & 0.5897 & 1.0448 & 46.84 & 26.08 & 53.55 & 3.1052 & \underline{90.13} \\
    \abbr-1B-Gen & \underline{0.3881} & \underline{0.5315} & \underline{0.5993} & \underline{1.0389} & 
    \underline{46.99} & \underline{26.46} & \underline{53.85} & \underline{3.1686} & 89.88 \\
    \abbr-1B-Spe & \textbf{0.3902} & \textbf{0.5333} & \textbf{0.6003} & \textbf{1.0347} & \textbf{47.85} & \textbf{27.17} & \textbf{54.35} & \textbf{3.1697} & \textbf{90.25} \\

    \midrule[1.5pt]
    Real & 0.6664 & 0.7773 & 0.8317 & 0.9599 & - & - & - & - & - \\
    Seq2Seq~\cite{seq2seq} & 0.3745 & 0.4473 & 0.4918 & 1.1357 & 51.74 & 27.02 & 45.40 & 1.3672 & 90.40 \\
    TM2T~\cite{tm2t} & \underline{0.5172} & \underline{0.5891} & \underline{0.6289} & \underline{1.0368} & \underline{60.63} & \underline{38.10} & \underline{54.13} & \underline{2.0841} & \textbf{93.47} \\
    \abbr-1B-Gen & 0.4141 & 0.4867 & 0.5240 & 1.1085 & 53.91 & 30.02 & 47.78 & 1.5811 & 91.29 \\
    \abbr-1B-Spe & \textbf{0.5249} & \textbf{0.6019} & \textbf{0.6419} & \textbf{1.0312} & \textbf{60.81} & \textbf{38.62} & \textbf{54.28} & \textbf{2.1205} & \underline{93.45} \\
    \bottomrule
    \end{tabular}
}
\label{tab:eval_m2t}
\end{table*}

\subsection{Multimodal Tokenization}
\label{sec:tokenization}
We convert all inputs (text, audio, music) and motion outputs into a single discrete vocabulary \(Z\).  Text is tokenized with a standard subword tokenizer.  All audio modalities (speech, music, ambient sounds) are encoded by a VQ–VAE~\cite{wavtokenizer} into tokens of the form \texttt{<|AUD\_x|>}, with each audio sequence delimited by \texttt{[AUDIO\_BOS]} and \texttt{[AUDIO\_EOS]}.  Motion for any number of agents is quantized via \vqname into tokens \texttt{<|MOT\_y|>}, wrapping each agent’s clip with \texttt{[MOT\_BOS]} and \texttt{[MOT\_EOS]}.  Agents are separated by a special \texttt{[AGENT\_SEP]} token.  Because our motion representation is global‐position–aware, these tokens implicitly encode inter-agent spatial relationships, allowing \name to support arbitrary group sizes without further architectural changes.

\subsection{Motion Message Formulation}
\label{sec:message}

Each task is represented as a single token sequence with three concise segments:
- \textbf{Instruction}: a brief natural‐language prompt (e.g., “Generate dance motion from this music”).  
- \textbf{Condition}: all required inputs—text, audio, motion, etc.—each enclosed by its modality’s BOS/EOS tokens.  To enable flexible, composite tasks (e.g.\ specifying duration in text-to-motion or appending style cues in music-to-dance), optional condition segments are stochastically inserted during training.  
- \textbf{Reply}: the desired output tokens, wrapped in the appropriate BOS/EOS markers.

At inference, the model autoregressively generates the reply segment conditioned on the instruction and any present conditions. Examples of different tasks and their corresponding motion message formats are presented in ~\tabref{tab:motion_message}.

We illustrate a single motion message that simultaneously conditions on multi‐agent motion, audio, and text as follow.

\begin{mdframed}
[backgroundcolor=gray!20,innerleftmargin=4pt,innerrightmargin=4pt,skipabove=\baselineskip]
\noindent
\textbf{Instruction \(\mathcal{I}\):} 
\texttt{\textbf{\textcolor{green}{[INSTR\_BOS]}}} Generate coordinated dance motion for two agents w.r.t the given music and caption. \texttt{\textbf{\textcolor{green}{[INSTR\_EOS]}}}\\
\textbf{Condition \(\mathcal{C}\):} 
\texttt{\textbf{\textcolor{green}{[COND\_BOS]}}}
\texttt{\textbf{\textcolor{blue}{[TEXT\_BOS]}}} Two dancers perform complementary sequences: one spins clockwise, the other leaps upward. \texttt{\textbf{\textcolor{blue}{[TEXT\_EOS]}}}
\texttt{\textbf{\textcolor{red}{[AUDIO\_BOS]}}} <AUD\_23> … <AUD\_47> \texttt{\textbf{\textcolor{red}{[AUDIO\_EOS]}}}
\texttt{\textbf{\textcolor{green}{[COND\_EOS]}}}\\
\textbf{Reply \(\mathcal{R}\):} 
\texttt{\textbf{\textcolor{green}{[REPLY\_BOS]}}}
\texttt{\textbf{\textcolor{purple}{[MOT\_BOS]}}} <MOT\_12> … <MOT\_240> 
\texttt{\textbf{\textcolor{gray}{[AGENT\_SEP]}}}<MOT\_26> … <MOT\_35> \texttt{\textbf{\textcolor{purple}{[MOT\_EOS]}}} 
\texttt{\textbf{\textcolor{green}{[REPLY\_EOS]}}}
\end{mdframed}

\subsection{Generalist Pretraining and Specialist Fine‑Tuning}
\label{sec:training}

Training follows three compact stages:  

1) \textbf{Generalist Pretraining:} generate diverse subtasks from MotionHub by pairing each clip’s metadata (text, audio, interaction tags) with various reply targets (e.g.\ text+motion \(\rightarrow\) audio, audio \(\rightarrow\) text+motion, unconditional motion) and train the Transformer on full sequences (instruction + condition + reply) via next‐token prediction to learn a unified motion prior.  

2) \textbf{Generalist SFT:} fine‐tune on (condition, instruction) \(\rightarrow\) reply pairs from the nine benchmarks (e.g.\ text‑to‑motion, motion‑to‑text), supervising only the reply tokens so the model adapts to task constraints while retaining cross‐modal versatility. 

3) \textbf{Specialist SFT:} further fine‐tune the generalist on individual domains, yielding specialists that push beyond the generalist to achieve or surpass SOTA.

\subsection{Decoder‐Only Transformer Architecture}
\label{sec:architecture}

We adopt a decoder‐only (GPT‐style) Transformer rather than the encoder–decoder designs used in prior motion LLMs (e.g.\ T5-based MotionGPT). This architecture treats all tokens—condition, instruction, and reply—uniformly, simplifying multimodal sequence modeling.  It also scales naturally to very large parameter counts and massive datasets, and it integrates seamlessly with existing autoregressive LLM pretraining frameworks~\cite{llama3,qwen,gemma} and pretrained weights.

Together, these design choices yield \name: a single, scalable framework for arbitrary cross‐modal generation and understanding across text, audio and motion, in both single‐ and multi‐agent settings.

\section{Experiments}

\begin{table*}
    \centering
\caption{Comparison results on AIST++ ~\cite{fact} and FineDance ~\cite{finedance} datasets. The empty columns of previous methods indicate that they can not handle the task.}
\label{tab:eval_d2m_m2d}
\centering
\adjustbox{width=0.95\linewidth}{
\begin{tabular}{lccccccccc}
\toprule
\multirow{2}{*}{Methods} & \multicolumn{3}{c}{Music-to-Dance on AIST++} & \multicolumn{3}{c}{Music-to-Dance on FineDance} & \multicolumn{3}{c}{Dance-to-Music on AIST++} \\ \cmidrule(lr){2-4} \cmidrule(lr){5-7} \cmidrule(lr){8-10} 
 & FID$_k \downarrow$ & Div$_k \uparrow$ & BAS $\uparrow$ & FID$_k \downarrow$ & Div$_k \uparrow$ & BAS $\uparrow$ & BCS$ \uparrow$ & BHS$ \uparrow$ & F1$\uparrow$ \\
\midrule
Real & 17.10 & 10.60 & 0.2374 & - & 16.61 & 0.2120 & - & - & -  \\ \specialrule{0em}{1pt}{1pt} \hline 

FACT \cite{fact} & 35.35 & 5.94 &  0.2209 & 113.38 & 3.36 & 0.1831    & -     & -     & -   \\ \specialrule{0em}{1pt}{1pt}
Bailando \cite{bailando}  &  \underline{28.16} & \underline{7.83} & 0.2332 & 82.81 & 7.74 & 0.2029 & - & - & -  \\

EDGE \cite{edge}  & 42.16 & 3.96 & \underline{0.2334}  & 94.34 & 8.13 & 0.2116 & - & - & - \\ 

Lodge \cite{lodge} & 37.09& 5.58 &  $ \textbf{0.2423} $  & \underline{\text{45.56}}  & \text{6.75}   & \textbf{0.2397}   &  -   & -  &  -    \\

Foley \cite{foley} & -  & - &  - & - & - & - &  96.4 & 41.0 &  57.5      \\

CMT \cite{cmt} & - & - & - & - & - & - & 97.1 & \text{46.2} &  \text{62.6}     \\

D2MGAN \cite{d2m-gan} & -  & -&  -& - & -    & -    & 95.6 & 88.7&  93.1    \\
CDCD \cite{cdcd} &- &- & - & - & - & - & 96.5 & 89.3   & \text{92.7}     \\

LORIS \cite{loris} &- &- & -  & -     & -    & -    & \underline{98.6} & 90.8 & \text{94.5} \\

M$^3$GPT~\cite{m3gpt}& $ \textbf{23.01}$ & 7.85  & $ \text{0.2261}$ & \textbf{\text{42.66}} & 8.24 & \underline{\text{0.2231}} & \text{94.3} & 94.0 & 95.0 \\

\midrule
\abbr-1B-Gen & 38.13 & \textbf{9.88} & 0.2206 & 56.22 & \textbf{15.20} & 0.2066 & 97.9 & \underline{97.3} & \underline{97.6} \\
\abbr-1B-Spe & 46.19 & 9.71 & 0.2307 & 55.20 & \underline{15.10} & 0.2063 & \textbf{99.5} & \textbf{98.6} & \textbf{99.0} \\
\quad + FlowVQ & 32.75 & \underline{9.87} & 0.2125 & 45.89 & 14.61 & 0.2162 & - & - & - \\

\bottomrule
\end{tabular}}%
\end{table*}

\begin{table}
\centering
\caption{Motion Prediction and In-between results. The gray color denotes \textbf{\textcolor{gray}{motion in-between}}} 
\adjustbox{width=0.95\linewidth}{
    \begin{tabular}{ccccc}
    \toprule[1.5pt]
    Methods  & FID\(\downarrow\) & Div\(\rightarrow\) & ADE\(\downarrow\) & FDE\(\downarrow\) \\
    \midrule[1pt]
    Real  & - & 17.4524 & - & - \\
    \midrule
    \multirow{2}{*}{MDM~\cite{mdm}}
      & 93.1619 & \textbf{17.4747} & 0.0319 & 0.0354 \\
      & \cellcolor{lightgray}86.8525 & \cellcolor{lightgray}\underline{17.3956} & \cellcolor{lightgray}0.0500 & \cellcolor{lightgray}- \\[1ex]
    \multirow{2}{*}{MotionGPT~\cite{motiongpt}}
      & 88.0120 & \underline{17.4275} & 0.0317 & 0.0356 \\
      & \cellcolor{lightgray}92.1697 & \cellcolor{lightgray}\textbf{17.4113} & \cellcolor{lightgray}0.0498 & \cellcolor{lightgray}- \\[1ex]
    \multirow{2}{*}{\abbr-1B-Gen}
      & 62.3713 & 17.3252 & 0.0334 & 0.0347 \\
      & \cellcolor{lightgray}62.5173 & \cellcolor{lightgray}17.2822 & \cellcolor{lightgray}\underline{0.0273} & \cellcolor{lightgray}- \\[1ex]
    \multirow{2}{*}{\abbr-1B-Spe}
      & \underline{58.5075} & 17.3264 & \underline{0.0288} & \underline{0.0306} \\
      & \cellcolor{lightgray}\underline{61.0623} & \cellcolor{lightgray}17.2890 & \cellcolor{lightgray}0.0280 & \cellcolor{lightgray}- \\[1ex]
    \multirow{2}{*}{\quad + \vqname}
      & \textbf{57.7554} & 17.3324 & \textbf{0.0282} & \textbf{0.0299} \\
      & \cellcolor{lightgray}\textbf{60.7917} & \cellcolor{lightgray}17.3544 & \cellcolor{lightgray}\textbf{0.0270} & \cellcolor{lightgray}- \\
    \bottomrule
    \end{tabular}
}
\label{tab:eval_m2m}
\end{table}

\begin{table*}
    \centering
    \caption{Accuracy comparison of different motion tokenizers in motion reconstruction tasks.}
    \adjustbox{width=0.9\textwidth}{
    \begin{tabular}{l|ccc|ccc|cc}
    
    \toprule 
     Metrics & \multicolumn{3}{c|}{MPJPE} & \multicolumn{3}{c|}{PA-MPJPE} & \multirow{2}{*}{\centering ADE} & \multirow{2}{*}{\centering FDE}\\
    Data Formats & all & body & hand & all & body & hand & & \\
    \midrule

    vanilla Motion VQ~\cite{motiongpt} & 0.1600 & 0.1351 & 0.1790 & 0.0891 & 0.0988 & 0.0817 & 0.0101 & 0.0126 \\

    Motion FSQ-VAE~\cite{scamo} & 0.1216 & 0.1194 & 0.1234 & 0.0780 & 0.0938 & 0.0652 & 0.0103 & 0.0116 \\

    RVQ + Hier Trans~\cite{momask,m3gpt,t2mhifigpt} & 0.1352 & 0.1179 & 0.1483 & 0.0784 & 0.0884 & 0.0708 & 0.0071 & 0.0118 \\
    \midrule
    \vqname w.o. Flow Matching Decoder & \underline{0.0655} & \underline{0.0629} & \underline{0.0676} & \underline{0.0417} & \underline{0.0494} & \underline{0.0355} & \underline{0.0056} & \underline{0.0081}
 \\
    \vqname(ours) & \textbf{0.0590} & \textbf{0.0557} & \textbf{0.0614} & \textbf{0.0394} & \textbf{0.0461} & \textbf{0.0343} & \textbf{0.0032} & \textbf{0.0045}
 \\

    \bottomrule
    \end{tabular}}
    \label{tab:eval_recons}
\end{table*}

\begin{figure*}
    \centering
    \includegraphics[width=0.9\linewidth]{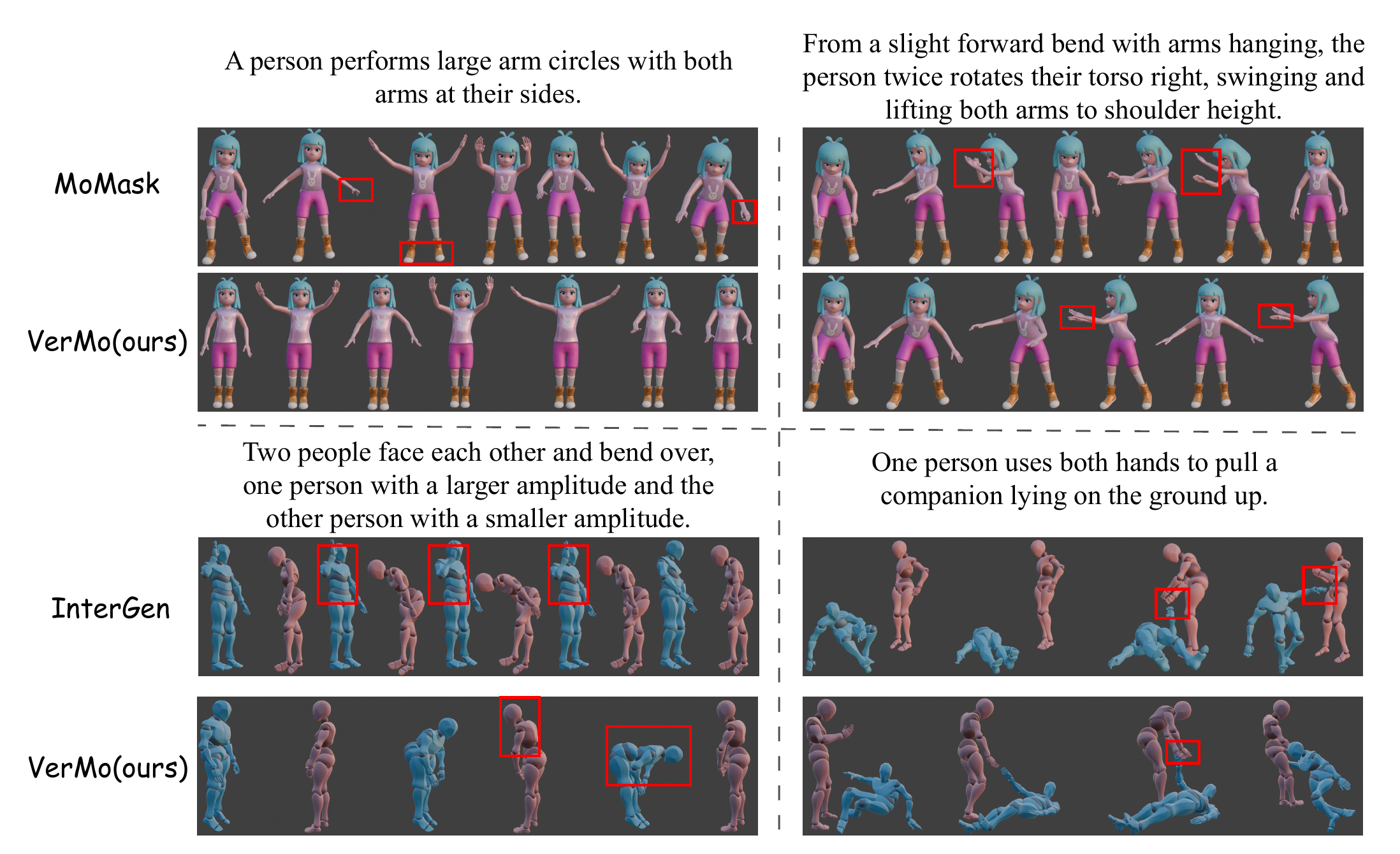}
    \caption{Qualitative comparison of \name and previous state‑of‑the‑art methods on single‑ and multi‑agent text‑to‑motion tasks.}   
    \label{fig:comp_t2m}
\end{figure*}

\begin{figure}
    \centering
    \includegraphics[width=0.95\linewidth]{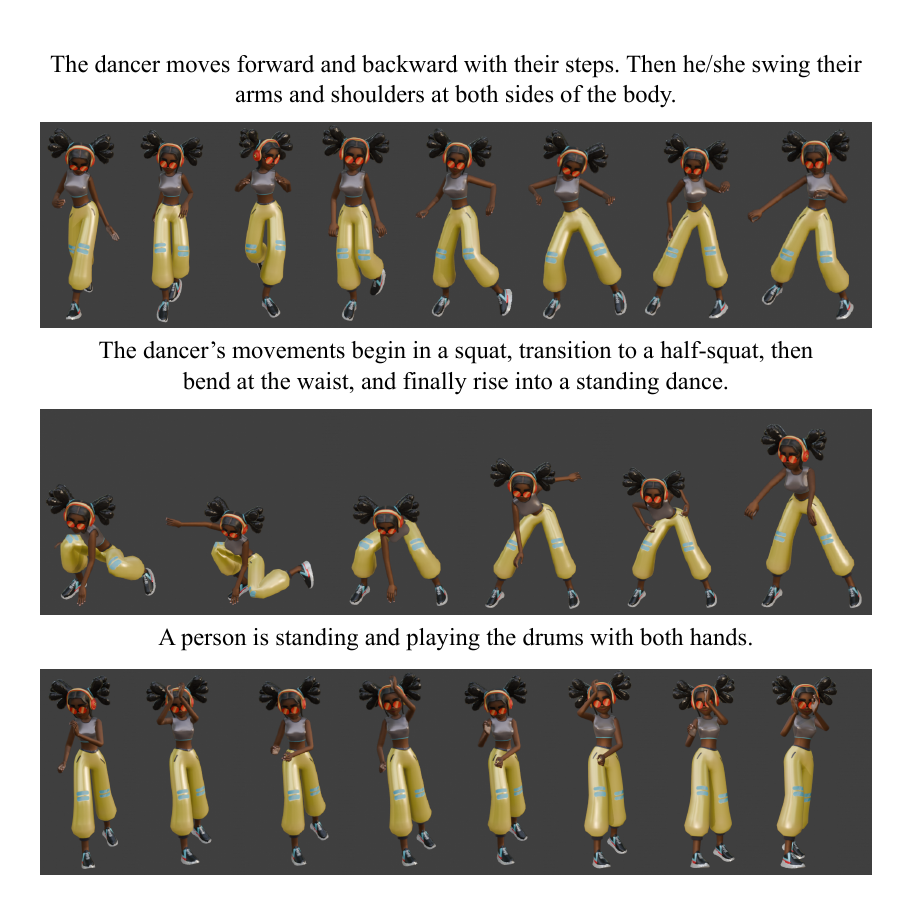}
    \caption{The motion generation results of \name driven jointly by audio and text.}   
    \label{fig:audio_text_dance}
\end{figure}

\begin{table}
    \centering
    \caption{Speech-to-Gesture evaluation on MotionHub}
    \adjustbox{width=0.95\linewidth}{
    \begin{tabular}{lcccc}
        \toprule[1.5pt]
        Method & FGD $\downarrow$ & BA $\rightarrow$ & L1Div$\uparrow$ \\
        \midrule
        Real & - & 0.2061 & 31.54 & \\
        CaMN~\cite{beat} & 65.18 & 0.1940 & 25.28 \\
        DiffStyleGesture~\cite{diffusestylegesture} & 58.20 & \underline{0.2037} & 29.68 \\
        MCM~\cite{mcm}  & \textbf{29.46} & 0.2399 & 31.80 \\
        \abbr-1B-Gen & 36.53 & 0.2089 & 38.28 \\
        \abbr-1B-Spe  & 38.66 & \underline{0.2085} & \textbf{40.43} \\
        \quad +FlowVQ & \underline{33.09} & \textbf{0.2046} & \underline{38.87}  \\
        \bottomrule[1.5pt]
    \end{tabular}
    }
    \label{tab:eval_s2g}
    
\end{table}
\begin{table}
    \centering
    \caption{Results of \name on T2M and M2T tasks with different initializations and architectures. Results in \textcolor{gray}{gray} cells correspond to M2T, while results in white cells correspond to T2M.}
    \adjustbox{width=\columnwidth}{
    \begin{tabular}{lccccc}
        \toprule
        Init & FID & \multicolumn{2}{c}{R Top 3} & \multicolumn{2}{c}{MM Dist} \\
        \midrule
        VM-1B-Random & 128.0712 & 0.2510 & \cellcolor{lightgray}0.1702 & 1.2695 & \cellcolor{lightgray}1.2966 \\
        VM-1B-LLaMA3.2 & 67.1204 & 0.4388 & \cellcolor{lightgray}0.4650 & 1.1608 & \cellcolor{lightgray}1.1461 \\
        VM-2B-Gemma2 & 69.1700 & 0.3793 & \cellcolor{lightgray}0.4625 & 1.1859 & \cellcolor{lightgray}1.1484 \\
        VM-3B-LLaMA3.2 & \underline{65.8164} & \underline{0.4463} & \cellcolor{lightgray}\underline{0.4697} & \underline{1.1544} & \cellcolor{lightgray}\underline{1.1455} \\
        VM-3B-QWen2.5 & 68.1545 & 0.4210 & \cellcolor{lightgray}0.4657 & 1.1676 & \cellcolor{lightgray}1.1458 \\
        VM-8B-LLaMA3.1 & \textbf{63.8843} & \textbf{0.4539} & \cellcolor{lightgray}\textbf{0.4842} & \textbf{1.1471} & \cellcolor{lightgray}\textbf{1.1365} \\
        \bottomrule
    \end{tabular}
    }
    \label{tab:abl_init}
\end{table}
\begin{table*}
    \centering
    \caption{Performance results of the \name-1B model on single-agent (white) and multi-agent (gray) text $\rightarrow$ motion and motion $\rightarrow$ text tasks under various training schemes.}
    \adjustbox{width=0.9\linewidth}{
    \begin{tabular}{lcccccccc}
        \toprule
        \multirow{2}{*}{Training Schemes} & \multicolumn{4}{c}{Text-to-Motion} & \multicolumn{4}{c}{Motion-to-Text}\\
        \cmidrule(lr){2-5} \cmidrule(lr){6-9}
        & FID $\downarrow$ & R-Top 3 $\uparrow$ & MMDist $\downarrow$ & Div $\rightarrow$ & MMDist $\downarrow$ & R-Top 3 $\uparrow$ & BLEU@4 $\uparrow$ & CIDEr-D $\uparrow$ \\
        \midrule
        \multirow{2}{*}{Real} & - & 0.7716 & 0.9834 & 17.4524 & 0.9834 & 0.7650 & - & - \\

        & \cellcolor{lightgray}- & \cellcolor{lightgray}0.8311 & \cellcolor{lightgray}0.9599 & \cellcolor{lightgray}23.3325 & \cellcolor{lightgray}0.9599 & \cellcolor{lightgray}0.8317 & \cellcolor{lightgray}- & \cellcolor{lightgray}- \\
        
        \multirow{2}{*}{Single Task Training} & 75.3696 & \underline{0.5071} & \underline{1.1181} & 17.1582 & 1.0449 & 0.5881 & 26.38 & 3.1316 \\
        & \cellcolor{lightgray}216.7339 & \cellcolor{lightgray}\underline{0.5444} & \cellcolor{lightgray}\underline{1.1419} & \cellcolor{lightgray}23.2039 & \cellcolor{lightgray}\underline{1.0379} & \cellcolor{lightgray}\underline{0.6250} & \cellcolor{lightgray}\underline{38.29} & \cellcolor{lightgray}\underline{2.1079} \\
        
        \multirow{2}{*}{Generalist Pretrain} & \textbf{63.4187} & 0.4407 & 1.1563 & \textbf{17.4337} & 1.0522 & 0.5800 & 25.33 & 2.9905 \\
        & \cellcolor{lightgray}160.5581 & \cellcolor{lightgray}0.4126 & \cellcolor{lightgray}1.2059 & \cellcolor{lightgray}\underline{23.3206} & \cellcolor{lightgray}1.1899 & \cellcolor{lightgray}0.3933 & \cellcolor{lightgray}23.74 & \cellcolor{lightgray}1.1095 \\
        
        \multirow{2}{*}{\quad + Generalist SFT} & \underline{63.4194} & 0.4889 & 1.1337 & \underline{17.4017} & \underline{1.0389} & \underline{0.5993} & \underline{26.46} & \underline{3.1686} \\
        & \cellcolor{lightgray}\underline{147.1982} & \cellcolor{lightgray}0.5051 & \cellcolor{lightgray}1.1571 & \cellcolor{lightgray}\textbf{23.3433} & \cellcolor{lightgray}1.1047 & \cellcolor{lightgray}0.5352 & \cellcolor{lightgray}30.13 & \cellcolor{lightgray}1.5693 \\
        
        \multirow{2}{*}{\quad \quad + Specialist SFT} & 64.7033 & \textbf{0.5304} & \textbf{1.0914} & 17.2965 & \textbf{1.0347} & \textbf{0.6003} & \textbf{27.17} & \textbf{3.1697} \\
        & \cellcolor{lightgray}\textbf{138.2251} & \cellcolor{lightgray}\textbf{0.6353} & \cellcolor{lightgray}\textbf{1.0939} & \cellcolor{lightgray}23.2228 & \cellcolor{lightgray}\textbf{1.0312} & \cellcolor{lightgray}\textbf{0.6419} & \cellcolor{lightgray}\textbf{38.62} & \cellcolor{lightgray}\textbf{2.1205} \\
        \bottomrule
    \end{tabular}
    }
    \label{tab:abl_train_stage_t2m_m2t}
\end{table*}
\begin{table*}
    \centering
\caption{Performance results of the \name-1B model on music $\leftrightarrow$ dance tasks under various training schemes.}
\label{tab:abl_train_stage_m2d_d2m}
\centering
\resizebox{0.9\textwidth}{!}{%
\begin{tabular}{lccccccccc}
\toprule
\multirow{2}{*}{Training Schemes} & \multicolumn{3}{c}{Music-to-Dance on AIST++} & \multicolumn{3}{c}{Music-to-Dance on FineDance} & \multicolumn{3}{c}{Dance-to-Music on AIST++} \\ \cmidrule(lr){2-4} \cmidrule(lr){5-7} \cmidrule(lr){8-10} \specialrule{0em}{1pt}{1pt}
 & FID$_k \downarrow$ & Div$_k \uparrow$ & BAS $\uparrow$ & FID$_k \downarrow$ & Div$_k \uparrow$ & BAS $\uparrow$ & BCS$ \uparrow$ & BHS$ \uparrow$ & F1$\uparrow$ \\ \specialrule{0em}{1pt}{1pt}
\midrule
Real & 17.10 & 10.60 & 0.2374 & - & 16.61 & 0.2120 & - & - & -  \\ 
Single Task Training & \underline{39.95} & \textbf{10.20}  & \textbf{0.2414} & 100.25 & 11.89 & \underline{0.2197} & 96.89 & \underline{97.88} & 97.38 \\
Generalist Pretrain & 47.77 & 4.97 & \underline{0.2337} & 142.18 & 13.27 & \textbf{0.2200} & 95.06 & 94.77 & 94.91 \\
\quad + Generalist SFT & \textbf{38.13} & \underline{9.88} & 0.2206 & \underline{56.22} & \textbf{15.20} & 0.2066 & \underline{97.88} & 97.27 & \underline{97.57} \\
\quad \quad + Specialist SFT & 46.19 & 9.71 & 0.2307 & \textbf{55.20} & \underline{15.10} & 0.2063 & \textbf{99.50} & \textbf{98.57} & \textbf{99.03} \\
\bottomrule
\end{tabular}}
\end{table*}

\subsection{Experiment Setup}  
We organize our experiments along three key dimensions:
\begin{itemize}[leftmargin=5mm, label=\normalfont\textbullet, itemsep=0.5ex, topsep=0.5ex, parsep=0ex]
  \item \textbf{Quantitative Benchmark Comparison:} numerical metrics on nine tasks (single‑/multi‑agent text \(\leftrightarrow\) motion, music \(\leftrightarrow\) dance, speech \(\rightarrow\) gesture, motion in‑between, prediction). In addition, we compare the reconstruction quality of \vqname with that of other motion tokenizers to assess their performance.All metric definitions and implementation details appear in ~\appsecref{sec:add_eval_metrics}.
  \item \textbf{Qualitative Comparison and Demonstrations:} visual comparisons with SOTA baselines and examples of \name generated under composite conditions.
  \item \textbf{Ablation Studies:} We conduct ablation studies from following perspectives:
  \textbf{Impact of Motion Tokenization:} We investigate the effect of using or omitting \vqname on motion synthesis tasks. \textbf{Training Strategy Analysis:} We evaluate how different training strategies influence the overall performance of \name. \textbf{Initialization and Scaling:} We analyze the impact of various initialization methods and scale settings on the behavior and stability of \name. \textbf{Ablation on Motion VQ-VAE Hyperparameters and Training Strategies:} \vqname consists of a Motion VQ-VAE and a Flow Matching Decoder. We conduct extensive ablation studies on the hyperparameters and training strategies of the Motion VQ-VAE component, with detailed results provided in ~\appsecref{sec:abl_motion_tokenizer}.
  
\end{itemize}

\subsection{Quantitative Benchmark Comparison}
\label{sec:quantitative}

\subsubsection{Text \(\leftrightarrow\) Motion}
\paragraph{Text-to-Motion Evaluation}  
Single-agent and multi-agent T2M evaluations employ the same set of metrics. R-Precision and MM-Distance assess the semantic alignment between generated motions and their corresponding captions, while FID measures the visual quality of the synthesized motions. As shown in the upper part of  ~\tabref{tab:eval_t2m}, in single-agent T2M, the \name generalist—pretrained on a mixture of tasks—acquires a strong motion prior and outperforms comparative baselines in motion quality. Further fine-tuning in the specialist stage yields additional gains in text–motion matching. As shown in the lower part of ~\tabref{tab:eval_t2m}, in multi-agent T2M, the generalist already surpasses several expert models; after specialist fine-tuning, \name achieves substantial improvements over all prior methods across both quality and alignment metrics.

\paragraph{Motion-to-Text Evaluation}  
For M2T, we employ R-Precision and MM-Distance to measure the semantic alignment between the generated captions and the input motion, and standard NLP metrics (e.g., BLEU, ROUGE, CIDEr, BERTScore) to assess consistency with ground-truth text candidates.  As shown in the upper half of ~\tabref{tab:eval_m2t}, the \name generalist already outperforms previous state-of-the-art in single-agent M2T, and subsequent specialist fine-tuning yields further improvements across all evaluation metrics.

\subsubsection{Motion In‑between and Prediction}
The evaluation results for the motion completion tasks (motion prediction and in-between) are presented in ~\tabref{tab:eval_m2m}. In the motion prediction task, the model is given the first 40\% of a sequence (past motion) and must generate the remaining 60\% (future motion).  In the in-betweening task, it is provided with the first and last 20\% of the sequence (past and future motion) and must interpolate the missing middle 60\% (middle motion). We evaluate generation quality and diversity using FID and Diversity—identical to the T2M setup—and measure temporal accuracy with ADE and FDE.  Because motion completion tasks constituted a large fraction of the generalist’s pretraining data, the generalist model already surpasses previous state-of-the-art performance on both prediction and in-betweening, and specialist fine-tuning yields no significant additional gains.  

\subsubsection{Music \(\leftrightarrow\) Dance}
For music-to-dance, we evaluate motion quality using Kinetic FID (\(\mathrm{FID}_k\)), diversity using \(\mathrm{Div}_k\), and rhythm alignment using Beat Alignment Score (BAS).  Following M\(^3\)GPT, we conduct separate evaluations on the AIST++ and FineDance subsets. As shown in ~\tabref{tab:eval_d2m_m2d}, thanks to large-scale motion pretraining, both the \name generalist and specialist achieve substantially higher \(\mathrm{Div}_k\) than prior domain-specific models, while delivering comparable BAS.  
In the dance-to-music direction, as illustrated in ~\tabref{tab:eval_d2m_m2d},\name’s strength is pronounced: the generalist alone surpasses all existing methods in rhythm matching, and this advantage further increases after specialist fine-tuning.  

\subsubsection{Speech \(\rightarrow\) Gesture}
We evaluate gesture quality using Fréchet Gesture Distance (FGD), rhythm alignment with Beat Alignment (BA), and diversity with L1Div.  As shown in ~\tabref{tab:eval_s2g}, thanks to the strong motion priors learned from multiple tasks, \name achieves significantly higher L1Div than previous methods, while matching or exceeding prior state-of-the-art in both FGD and BA.  

\subsubsection{Motion Tokenizer Reconstruction}

We measure motion reconstruction error in the original coordinate space using Mean Per Joint Position Error (MPJPE), and compute Procrustes-aligned MPJPE (PA-MPJPE) by discounting global root translation.  Root trajectory accuracy is evaluated with Average Displacement Error (ADE) and Final Displacement Error (FDE), corresponding to the mean and final‐frame root position errors, respectively.  

Prior Motion VQ-VAE tokenizers~\cite{t2mgpt,motiongpt,scamo} achieve only modest reconstruction quality.  Methods such as ~\cite{m3gpt,t2mhifigpt} employ RVQ to predict multiple residual tokens per time step via a hierarchical Transformer, but accurately forecasting this stack of codes remains challenging, limiting reconstruction fidelity.  In contrast, we first apply architectural and hyperparameter optimizations to the standard Motion VQ-VAE backbone (named \vqname w.o. Flow Matching Decoder) and then introduce a Flow Matching Transformer decoder.  This two‐stage refinement yields substantially lower MPJPE, PA-MPJPE, ADE, and FDE, outperforming all previous motion tokenizers by a wide margin.

\subsection{Qualitative Results}
\label{sec:qualitative}
We visualize representative samples across tasks, comparing \name’s outputs to SOTA baselines.  Composite‑condition cases (e.g.\ text+audio \(\rightarrow\) motion, motion \(\rightarrow\) text+audio) demonstrate \name’s ability to flexibly fuse modalities.  Detailed qualitative results are in ~\appsecref{sec:demo}.

\subsubsection{Qualitative Comparison}

In addition to the quantitative evaluations above, we also performed qualitative comparisons against previous state‑of‑the‑art methods.

As shown in the upper part of ~\figref{fig:comp_t2m}, we perform a qualitative comparison between MoMask and \name on the single‑agent text‑to‑motion task. As highlighted by the red boxes, MoMask’s outputs exhibit hand mesh interpenetration and twisting, as well as misaligned foot placement, whereas \name delivers markedly higher quality and finer anatomical detail.

In the lower part of ~\figref{fig:comp_t2m}, we compare \name with InterGen on multi‑agent text‑to‑motion generation. As highlighted by the red boxes, InterGen fails to capture the nuanced instruction “one agent bows deeply, the other bows slightly,” whereas \name accurately renders this distinction. Moreover, for fine-grained interactive actions such as “pulling,” \name demonstrates notably more precise hand movements and inter‑agent coordination.

We evaluate \name's capability in handling composite conditional tasks, which are challenging or unsupported in previous work. As shown in ~\figref{fig:audio_text_dance}, our generalist model demonstrates strong performance under such complex settings. \name generates motions that not only align well with the audio but also accurately reflect the composite textual descriptions.

\subsection{Ablation Study}

\subsubsection{Ablation Study on \vqname}
As shown in ~\tabref{tab:eval_recons}, we first quantify the incremental gains that \vqname brings to motion reconstruction, demonstrating a marked improvement in reconstruction fidelity. While tokenizers often face an “optimization dilemma”~\cite{vavae} — where enhancements in reconstruction do not necessarily translate to better generative performance—we evaluate both our generalist and specialist models with and without \vqname across motion generation benchmarks (see ~\tabref{tab:eval_t2m}, ~\tabref{tab:eval_d2m_m2d}, ~\tabref{tab:eval_s2g}, ~\tabref{tab:eval_m2m}.) In every scenario, integrating \vqname yields immediate, plug-and-play gains in generation quality.

\subsubsection{Ablation Study on Training Stages}

We compare several training regimes to assess their impact on \name across different tasks. Specifically, we evaluate: (1) specialist fine-tuning only; (2) generalist pretraining only; (3) generalist pretraining followed by specialist fine-tuning; and (4) a three‐stage pipeline of generalist pretraining, generalist fine-tuning, and specialist fine-tuning. ~\tabref{tab:abl_train_stage_t2m_m2t} reports results on single‐agent and multi‐agent text→motion (T2M) and motion→text (M2T) benchmarks. For data‐rich single‐agent T2M and M2T tasks, specialist fine-tuning alone and generalist fine-tuning after pretraining yield comparable performance. However, on data‐scarce multi‐agent T2M and M2T tasks, the motion priors learned during generalist pretraining are essential, significantly boosting the downstream specialist’s performance.

\subsubsection{Ablation Study on Transformer Initialization}
\label{sec:abl_init} 
The autoregressive Transformer module of \name can theoretically use any type of decoder-only Transformer. Leveraging pretrained LLM models for initialization allows the autoregressive Transformer to inherit the powerful semantic understanding and text generation capabilities of language models. The goal of this experiment is to select the most suitable decoder architecture and pretrained weights as the structure and initialization parameters for \name. Training \name fully requires three stages: generalist pretraining, generalist SFT, and specialist SFT, with generalist training being particularly costly. Therefore, for this experiment, we only conduct joint training on the T2M and M2T tasks, which reflect the model's capabilities for both motion synthesis and comprehension. To reduce training costs and ensure fair comparison, each model is trained for 10 epochs in an LLM pretraining manner on both tasks. As shown in ~\tabref{tab:abl_init}, VM-1B-Random (trained from scratch) performs significantly worse. We leverage three widely adopted LLM architectures—Qianwen~\cite{qwen}, Gemma~\cite{gemma}, and LLaMA—and further evaluate the impact of model scale by instantiating LLaMA at three sizes: 3.2-1B, 3.2-3B, and 3.1-8B. As shown in ~\tabref{tab:abl_init}, increases in parameter count yield consistent gains in both synthesis and comprehension capabilities of \name. Balancing performance and computational efficiency, we ultimately select LLaMA-3.2-1B as \name’s initialization backbone, despite its not being the top performer. We note that with greater compute resources and a more powerful LLM initialization, \name’s performance stands to improve significantly.

\section{Conclusion}

We have introduced \name, the first unified multimodal Motion LLM that natively handles both single- and multi-agent scenarios and enables cross-modal conversion among text, audio, and motion.  Central to \name is \vqname, a novel Motion Tokenizer that replaces the standard VQ–VAE decoder with a Flow Matching Transformer, yielding marked gains in reconstruction fidelity and downstream synthesis.  

To fuel large-scale pretraining and support a variety of motion-domain tasks as benchmarks, we present MotionHub — the largest unified motion dataset to date—with over 350 K motion sequences (approximately 600 hours in total) annotated with captions, audio, transcripts, and interaction data, and split into nine standardized benchmarks.

Finally, our three-stage training pipeline produces a robust generalist model that surpasses many task-specific baselines and, with targeted fine-tuning, specialist variants that achieve state-of-the-art performance on seven of nine tasks.  Together, these advances lay a scalable foundation for future work in multimodal motion understanding and generation.

\appendix

{
\bibliographystyle{ACM-Reference-Format}
\bibliography{main}

}
\newpage

\clearpage
\setcounter{page}{1}
\begin{figure*}
    \centering
    \includegraphics[width=0.9\linewidth]{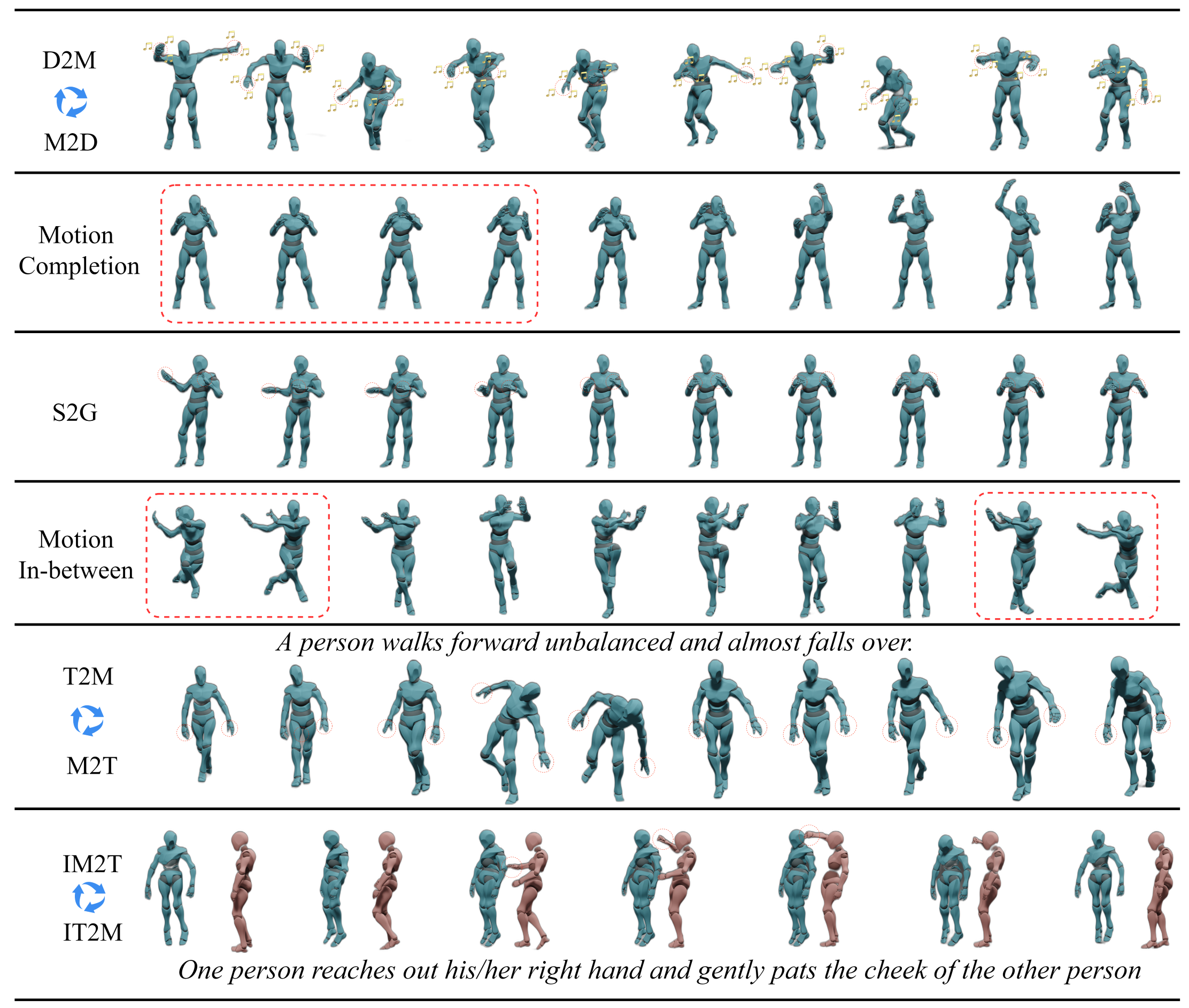}
    \caption{Samples from the different tasks in our constructed MotionHub.}   
    \label{fig:motionhub} 
\end{figure*}

\begin{figure}[htbp]
    \centering
    \includegraphics[width=0.95\linewidth]{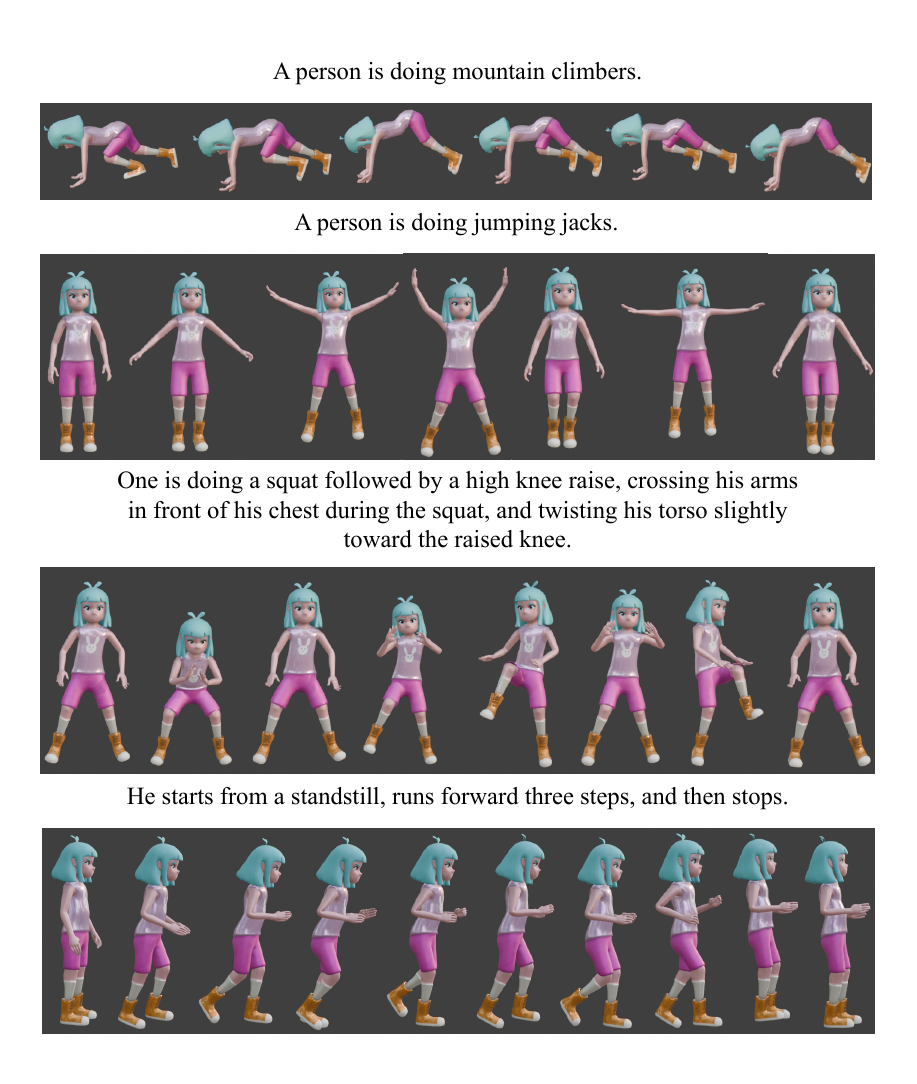}
    \caption{Single-Person Text-to-Motion visualization samples.}   
    \label{fig:demo_t2m}
\end{figure}

\begin{figure}[htbp]
    \centering
    \includegraphics[width=0.95\linewidth]{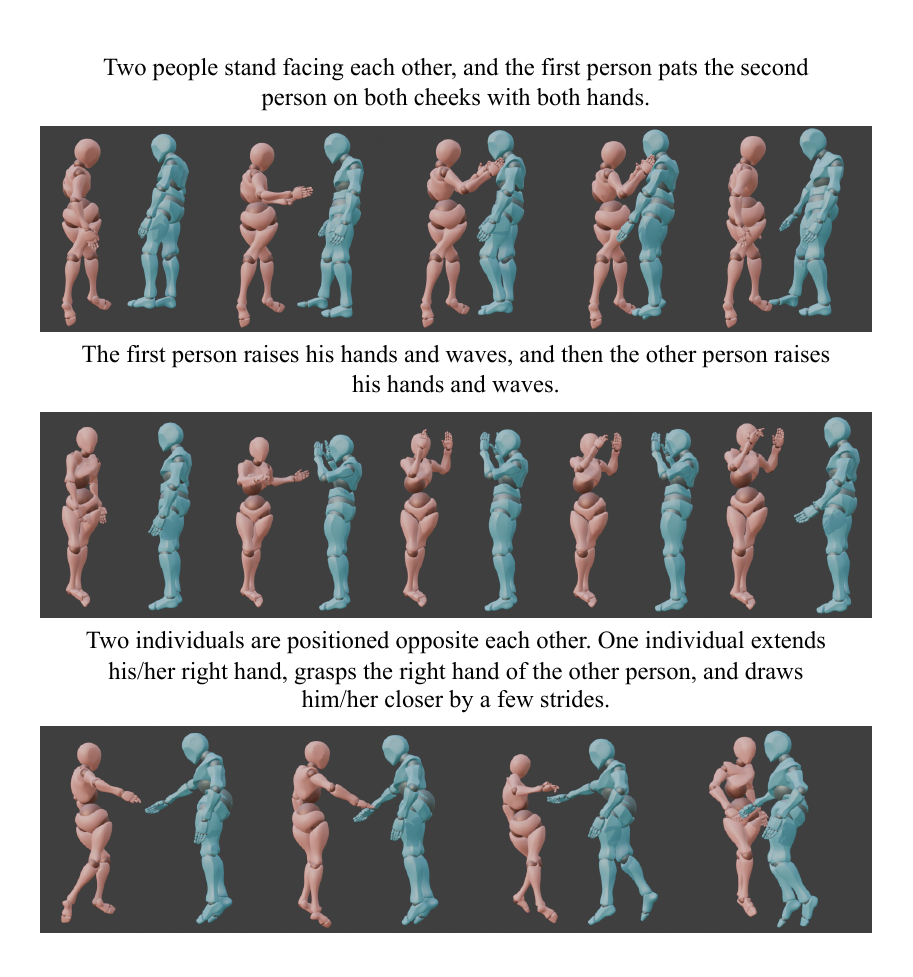}
    \caption{Dual-agent Text-to-Motion visualization samples.}   
    \label{fig:demo_t2m_2p}
\end{figure}

\begin{figure}[htbp]
    \centering
    \includegraphics[width=0.95\linewidth]{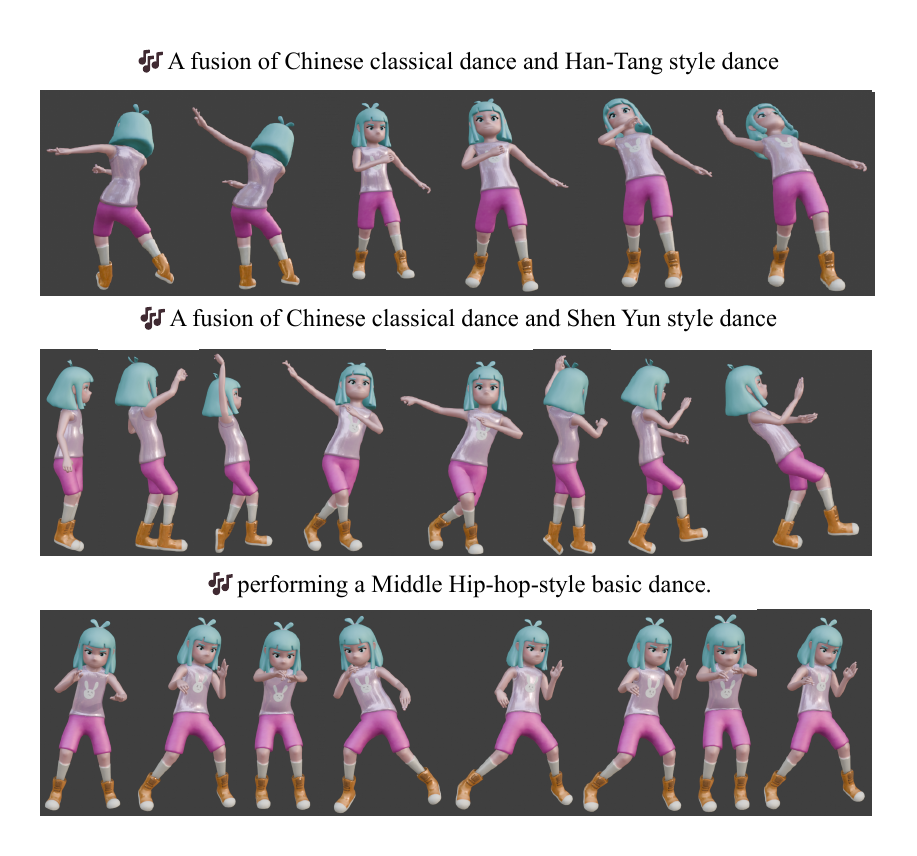}
    \caption{Music-to-Dance visualization samples.}   
    \label{fig:demo_m2d}
\end{figure}

\begin{figure}
    \centering
    \includegraphics[width=0.95\linewidth]{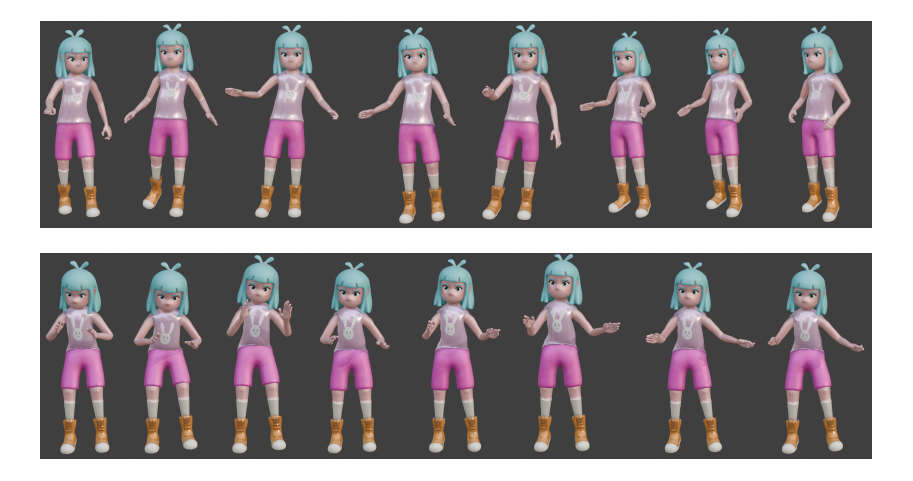}
    \caption{Speech-to-Gesture visualization samples.}   
    \label{fig:demo_s2g}
\end{figure}

\begin{figure}
    \centering
    \includegraphics[width=0.95\linewidth]{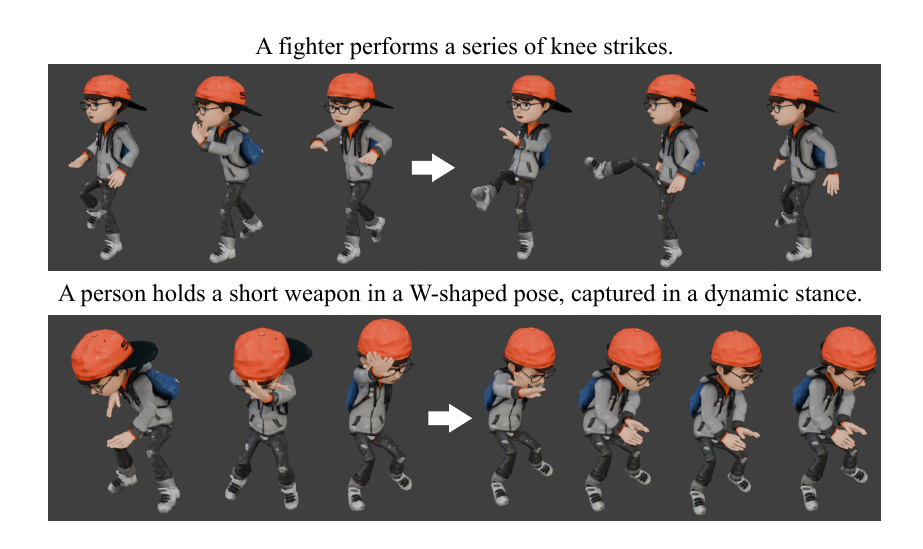}
    \caption{Motion Prediction visualization samples. The arrow indicates the movement direction. }   
    \label{fig:demo_pred}
\end{figure}

\begin{figure}
    \centering
    \includegraphics[width=0.95\linewidth]{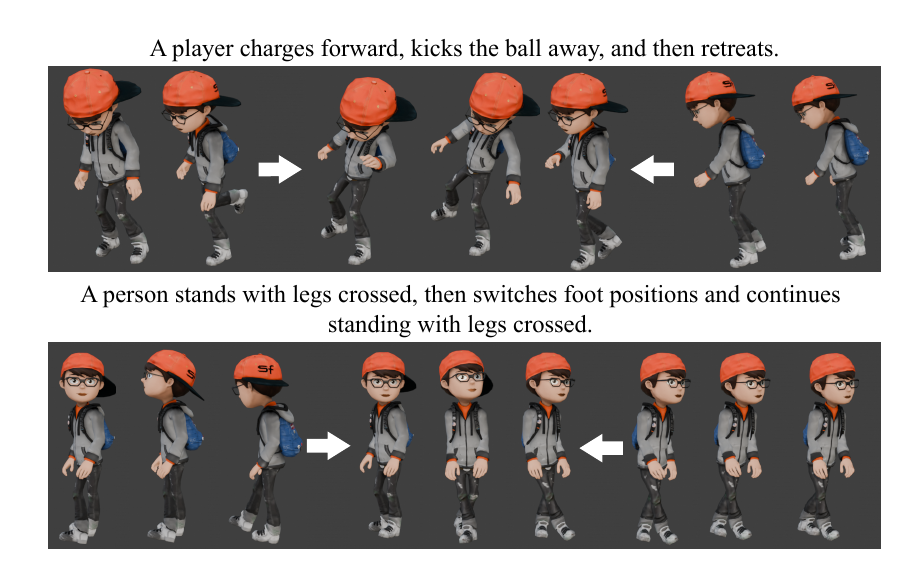}
    \caption{Motion Inbetween visualization samples. The motion between the two arrows are predictd by \name, while the others are given as condition.}   
    \label{fig:demo_inbetween}
\end{figure}

\section{Implementation Details of \name}

Our implementation is built on the PyTorch 2~\cite{pytorch} and MMEngine~\cite{mmengine} frameworks. All models, including \name and baseline methods, are trained and evaluated using the Fully Sharded Data Parallel (FSDP) strategy with bfloat16 precision for both training and inference. Except for T5, all Transformer-based models are accelerated using FlashAttention-2. The codebase, named MMotion, not only supports \name but also provides one-click training and evaluation for a series of prior works in the motion domain on the MotionHub dataset. Additionally, it includes utility tools for motion data processing, visualization, and other related functionalities.

\subsection{Multimodal Tokenizer}
The multimodal tokenizer consists of three components: the motion tokenizer (\vqname), the audio tokenizer, and the text tokenizer. The motion tokenizer and audio tokenizer are based on VQ-VAE, containing 4,375 and 4,096 tokens, respectively. These tokens are incorporated into the text tokenizer in the form of symbolic text tokens. For example, the 4,375 motion tokens are represented as \texttt{<|MOTION\_1|>} to \texttt{<|MOTION\_4375|>}. In addition to the motion and audio tokens, 29 additional special tokens are added to the vocabulary to indicate the start and end of different modalities or various segments within the motion message.
\noindent \textbf{\vqname}
\vqname includes a VQ-VAE architecture consisting of an encoder, FSQ-based quantizer, and decoder, and a Transformer-based Flow Mathcing Decoder.

The VQ-VAE encoder and decoder employ mirrored structures with 3 hierarchical layers, each containing 3 ResNet blocks. The encoder applies 3 ResNet blocks followed by a downsampling convolution (kernel=4, stride=2) in each layer, with an intermediate layer near the quantizer adding 2 additional ResNet blocks, and uses 3\(\times\)3 convolutions for input/output layers. The decoder performs nearest-neighbor upsampling followed by a 3\(\times\)3 convolution and 3 ResNet blocks in each layer, with the final output layer stacking a 3\(\times\)3 convolution, ReLU activation, and another 3\(\times\)3 convolution. The architecture features a uniform channel width of 512 across layers, a temporal downsampling rate of 4$\times$4, ReLU activations without normalization layers, and no dilated convolutions. The quantizer implements FSQ with a codebook size of 4,375 derived from discrete levels [7,5,5,5,5], and all convolutional operations use kernel size 3 except for the down/upsampling layers.

Flow Matching Transformer Decoder is composed of eight sequential Transformer blocks. Each block consists of, in order, a cross‐attention layer, a self‐attention layer, and a feed‐forward (FFN) layer. Both attention layers are multi‐head with four heads of 128 channels each, and the FFN hidden dimension is 1024. In the cross‐attention, keys and values are formed by concatenating the timestep embedding with the quantizer’s output embeddings z.

For the flow matching scheduler, we train for 1,000 timesteps and perform 28 inference steps with a shift parameter of 3.0, using the FlowMatchEulerDiscreteScheduler implementation from the Diffusers~\cite{diffusers} library.

Motion VQ++ is trained on a single NVIDIA A800-80G GPU for 1,000 epochs on the MotionHub dataset, requiring approximately 48 hours. The training configuration includes a batch size of 6,400, the AdamW optimizer with $\beta_1=0.9$, $\beta_2=0.99$, zero weight decay, and a learning rate of $2 \times 10^{-4}$. During training, all motion samples are randomly cropped to a maximum of 256 frames (at 30 fps), and sequences within each batch are padded with zeros to match the length of the longest sequence in the batch. The loss function employs smooth L1 loss without commit loss, and for datasets lacking hand data (e.g., SMPL-based datasets), hand-related losses are excluded. Motion data is represented as 3D global positions (156-dimensional) and standardized similarly to HumanML3D, with details provided in~\appsecref{sec:abl_standardization}.

\noindent \textbf{Audio Tokenizer}
\name is specifically designed for audio-related tasks, encompassing both vocal and musical signals. Given the inherently long temporal nature of audio signals, conventional audio tokenizers typically suffer from low compression rates and rely on multiple codebooks, resulting in audio tokens that cannot be processed equivalently with text and motion modality tokens. To address this, we leverage Wavtokenizer~\cite{wavtokenizer}, a unified audio tokenization framework that supports diverse audio types (e.g., music, speech) while employing a single-codebook architecture with significantly higher temporal compression.

We utilize the pre-trained WavTokenizer-Large-Unify model for tokenizing both music and speech audio. All audio inputs are resampled to 24 kHz, with each second of audio compressed into 40 tokens. The tokenizer employs a codebook size of 4,096.

\noindent \textbf{Text Tokenizer}
Pre-trained LLM tokenizer as a text tokenizer has demonstrated sufficient reliability through extensive empirical validation. The text tokenizer handles the tokenization and detokenization of textual conditions (including captions, motion duration specifications, and participant counts), task instructions, and special symbolic markers.
We employ the pre-trained LLaMA-3 tokenizer as the text tokenizer, utilizing the Byte-Pair Encoding (BPE) algorithm. The original vocabulary size is 128,256.

\subsection{Autoregressive Transformer}

The model architecture is implemented based on the following configuration: it uses a LLaMA-style transformer with 16 hidden layers, each containing 32 attention heads and a head dimension of 64. The hidden size is set to 2,048, with an intermediate size of 8,192 for the feed-forward network. The model employs SiLU (Swish-1) as the hidden activation function and does not use biases in the attention or MLP layers. Layer normalization is applied with an epsilon value of $1 \times 10^{-5}$. The attention mechanism does not include bias terms or dropout, and the initializer range is set to 0.02. The model supports a maximum sequence length of 131,072 tokens through RoPE (Rotary Positional Embeddings) scaling, with a scaling factor of 32.0, a base frequency factor of 500,000, and high/low frequency factors of 4.0 and 1.0, respectively.

The Autoregressive Transformer undergoes joint pretraining on the entire MotionHub dataset across all tasks to obtain the \name-1B-Gen-pre model. Training is conducted on 8 NVIDIA A100-80G GPUs for 30 epochs with a batch size of 16 per GPU. The optimizer is AdamW with a learning rate of $1 \times 10^{-5}$, $\beta_1=0.9$, $\beta_2=0.99$, and zero weight decay. This stage takes around 48 hours. At this stage, all tokens (excluding padding tokens), including conditions and instructions, are required to participate in the next-token prediction task, and the corresponding loss (L1) is computed. In the next stage, the model continues training on the entire MotionHub dataset across all tasks but shifts to supervised fine-tuning (SFT), where only reply tokens are used to compute the loss. This stage consists of 30 epochs, requiring approximately 96 hours on an 8-card A100-80G machine, with all other training parameters remaining unchanged. The resulting model is denoted as \name-1B-Gen-SFT. Finally, we perform task-specific supervised fine-tuning (SFT) on the datasets corresponding to the 9 benchmark tasks. Each task is trained for 100 epochs on a single A100-80G GPU. The training time for Single-Person T2M, M2T, and M2M tasks is approximately 48 hours, while S2G, M2D, and D2M tasks require about 24 hours. Due to the smaller dataset sizes, Multi-Person T2M and M2T tasks are completed in approximately 10 hours.

\section{Ablation Study on Motion Tokenizer}
\label{sec:abl_motion_tokenizer}

This section presents systematic ablation studies across the motion tokenization pipeline, evolving from vanilla Motion VQ~\cite{motiongpt,tm2t,m3gpt} to the \vqname Tokenizer, with the objective of elucidating the role of each constituent element in the tokenization process.


\begin{table*}
    \centering
    \caption{Impact of motion representations on vanilla Motion VQ reconstruction quality}
    \adjustbox{width=0.9\linewidth}{
    \begin{tabular}{l|ccc|ccc|cc}
    
    \toprule 
     Metrics & \multicolumn{3}{c|}{MPJPE} & \multicolumn{3}{c|}{PA-MPJPE} & \multirow{2}{*}{\centering ADE} & \multirow{2}{*}{\centering FDE}\\
    Data Formats & all & body & hand & all & body & hand & & \\
    \midrule

    pos & 0.1835 & 0.1810 & 0.1855 & 0.0829 & 0.0964 & 0.0720 & 0.0208 & 0.0200 \\

    pos + vel & 0.1914 & 0.2004 & 0.2031 & 0.0658 & 0.0821 & 0.0592 & 0.0300 & 0.0281 \\

    pos + rot & 0.2907 & 0.2821 & 0.2976 & 0.0799 & 0.0925 & 0.0697 & 0.0405 & 0.0403 \\

    pos + vel + rot & 0.2930 & 0.2825 & 0.3015 & 0.0711 & 0.0834 & 0.0611 & 0.0454 & 0.0466 \\
    
    InterHuman & 0.2708 & 0.2594 & 0.2800 & 0.0738 & 0.0844 & 0.0651 & 0.0398 & 0.0410 \\
    \bottomrule
    \end{tabular}
    }
    \label{tab:abl_rep}
\end{table*}

\subsection{The Impact of Different Motion Representations}
Previous work predominantly employs HumanML3D~\cite{t2m} or Humantomato~\cite{humantomato} as motion representations. Both methods discard the global position information of agents, making them unsuitable for multi-person motion generation. InterHuman~\cite{intergen} introduces a composite motion representation that incorporates coordinates, velocity, rotation, foot contact, and other information. We deconstruct this representation to evaluate the influence of each component on the performance of the motion tokenizer. As shown in ~\tabref{tab:abl_rep}, we find that using only position as the motion representation is the simplest and most effective approach.

\begin{table*}
    \centering
    \caption{Comparison of motion reconstruction performance under different combinations of motion representations and standardization methods.}
    \adjustbox{width=0.9\linewidth}{
    \begin{tabular}{l|ccc|ccc|cc}
    
    \toprule 
     Metrics & \multicolumn{3}{c|}{MPJPE} & \multicolumn{3}{c|}{PA-MPJPE} & \multirow{2}{*}{\centering ADE} & \multirow{2}{*}{\centering FDE}\\
    Data Formats & all & body & hand & all & body & hand & & \\
    \midrule

    pos & 0.1835 & 0.1810 & 0.1855 & 0.0829 & 0.0964 & 0.0720 & 0.0208 & 0.0200 \\

    pos + standard & 0.2197 & 0.2110 & 0.2266 & 0.0804 & 0.0925 & 0.0707 & 0.0317 & 0.0288 \\

    pos + avg-std & 0.1723 & 0.1645 & 0.1786 & 0.0776 & 0.0896 & 0.0680 & 0.0215 & 0.0211 \\

    pos + min-max & 0.2119 & 0.2119 & 0.2119 & 0.1279 & 0.1481 & 0.1115 & 0.0250 & 0.0226 \\
    
    InterHuman & 0.2708 & 0.2594 & 0.2800 & 0.0738 & 0.0844 & 0.0651 & 0.0398 & 0.0410 \\

    InterHuman + avg-std & 0.1710 & 0.1649 & 0.1759 & 0.0712 & 0.0832 & 0.0616 & 0.0210 & 0.0205 \\

    \bottomrule
    \end{tabular}
    }
    \label{tab:abl_standardization}
\end{table*}

\subsection{Impact of Different standardization Methods} 
\label{sec:abl_standardization}
HumanML3D employs a standardization method where the mean ($\mu$) and standard deviation ($\sigma$) are computed, 
and the standard deviations across channels for each modality (postion, velocity, rotation, foot contact, etc.) 
are averaged. We compare four strategies:
\begin{itemize}
  \item \textbf{No standardization}:
    \begin{equation}
      x_{\text{norm}} = x
    \end{equation}

  \item \textbf{Standard standardization(standard)}:
    \begin{equation}
      x_{\text{norm}} = \frac{x - \mu}{\sigma}
    \end{equation}

  \item \textbf{HumanML3D-style (avg-std)}:
    For a modality with $C$ channels, compute a shared $\sigma_{\text{avg}}$:
    \begin{equation}
      \sigma_{\text{avg}} = \frac{1}{C}\sum_{c=1}^C \sigma_c
    \end{equation}
    Then apply:
    \begin{equation}
      x_{\text{norm}} = \frac{x - \mu}{\sigma_{\text{avg}}}
    \end{equation}

  \item \textbf{Min-Max scaling(min-max)}:
    \begin{equation}
      x_{\text{norm}} = \frac{x - x_{\min}}{x_{\max} - x_{\min}}
    \end{equation}
\end{itemize}
As shown in Table~\ref{tab:abl_standardization}, 
the avg-std method improves reconstruction metrics for both position-based and InterHuman representations, 
with InterHuman achieving slightly better results than position-based. 
However, balancing simplicity and effectiveness, 
we ultimately adopt the \textbf{position + avg-std} combination as our motion representation and standardization strategy.

\begin{table*}
    \centering
    \caption{Reconstruction metrics of Motion VQ-VAE with different depths.}
    \adjustbox{width=0.9\linewidth}{
    \begin{tabular}{l|ccc|ccc|cc}
    
    \toprule 
     Metrics & \multicolumn{3}{c|}{MPJPE} & \multicolumn{3}{c|}{PA-MPJPE} & \multirow{2}{*}{\centering ADE} & \multirow{2}{*}{\centering FDE}\\
    Dilation & all & body & hand & all & body & hand & & \\
    \midrule
    dilation=1 & 0.1671 & 0.1460 & 0.1840 & 0.0865 & 0.1014 & 0.0745 & 0.0127 & 0.0187 \\
    dilation=3 & 0.1702 & 0.1498 & 0.1873 & 0.0868 & 0.1018 & 0.0748 & 0.0141 & 0.0192 \\
    growth-dilated & 0.1888 & 0.1768 & 0.1984 & 0.0874 & 0.1023 & 0.0754 & 0.0192 & 0.0205 \\

    \bottomrule
    \end{tabular}
    }
    \label{tab:abl_dilation}
\end{table*}

\subsection{Ablation on Dilation Strategy}
\label{subsec:dilation}
Vanilla Motion VQ employs a progressive dilation strategy in its convolutional layers: 
for the $i$-th last ResNet block in each encoder and decoder layer, 
the dilation rate is set to $3^i$ (i.e., 1, 3, 9). 
We conduct an ablation study to compare three configurations:
\begin{itemize}
  \item \textbf{No dilation}: Standard convolutions without dilation.
  \item \textbf{Constant dilation}: Fixed dilation rate of 3 across all layers.
  \item \textbf{Progressive dilation (growth-dilated)}: Dilation rate increases as $3^i$.
\end{itemize}
As shown in Table~\ref{tab:abl_dilation}, 
the \textbf{no dilation} configuration achieves the best reconstruction performance. 
We hypothesize that motion reconstruction tasks do not require excessively large receptive fields, 
making standard convolutions more effective than dilated variants.

\begin{table*}
    \centering
    \caption{Comparison between different order of downsample and ResNet blocks}
    \adjustbox{width=0.9\linewidth}{
    \begin{tabular}{l|ccc|ccc|cc}
    
    \toprule 
     \multirow{2}{*}{\centering Pre} & \multicolumn{3}{c|}{MPJPE} & \multicolumn{3}{c|}{PA-MPJPE} & \multirow{2}{*}{\centering ADE} & \multirow{2}{*}{\centering FDE}\\
     & all & body & hand & all & body & hand & & \\
    \midrule

    \checkmark & 0.1818 & 0.1651 & 0.1954 & 0.0866 & 0.1000 & 0.0757 & 0.0183 & 0.0221 \\ 

    $\times$ & 0.1724 & 0.1548 & 0.1865 & 0.0859 & 0.0996 & 0.0748 & 0.0162 & 0.0193 \\
    
    \bottomrule
    \end{tabular}
    }
    \label{tab:abl_down_order}
\end{table*}

\subsection{Relative Position of Downsampling Layers and ResNet Blocks}
\label{subsec:down_order}

Vanilla Motion VQ adopts a \textbf{pre-downsample} strategy: 
in the encoder, after the input layer (a single convolutional layer followed by an activation function), 
each encoder layer first performs downsampling and then uses ResNet blocks to extract features. 
The advantage of this approach is that ResNet blocks operate on shorter temporal sequences, 
reducing computational cost. 
However, the downside is that failing to extract features at high temporal resolution 
may lead to suboptimal reconstruction quality. 

We investigate the position of downsampling relative to ResNet blocks as an experimental variable. 
As shown in ~\tabref{tab:abl_down_order}, 
placing downsampling \textbf{after} ResNet blocks ($\times$) 
outperforms placing it \textbf{before} (\checkmark). 
This suggests that high-resolution feature extraction is critical for achieving optimal reconstruction quality, 
even at the cost of increased computational complexity.

\begin{table*}
    \centering
    \caption{Results of vanilla Motion VQ with different quantizers with different codebook sizes.}
    \adjustbox{width=0.9\linewidth}{
    \begin{tabular}{ll|ccc|ccc|cc}
    
    \toprule 
     \multirow{2}{*}{\centering Size} & \multirow{2}{*}{\centering Quantizer} & \multicolumn{3}{c|}{MPJPE} & \multicolumn{3}{c|}{PA-MPJPE} & \multirow{2}{*}{\centering ADE} & \multirow{2}{*}{\centering FDE}\\
     & & all & body & hand & all & body & hand & & \\
     \midrule

     \multirow{5}{*}{\centering \(2^{9}\)} & VQ & 0.1888 & 0.1768 & 0.1984 & 0.0874 & 0.1023 & 0.0754 & 0.0192 & 0.0205 \\
    & FSQ & 0.2267 & 0.1937 & 0.2532 & 0.1303 & 0.1546 & 0.1109 & 0.0190 & 0.0221 \\
    & LFQ & 0.3186 & 0.2759 & 0.3529 & 0.1282 & 0.1473 & 0.1129 & 0.0298 & 0.0369 \\
    & RVQ(3) & 0.0599 & 0.0574 & 0.0619 & 0.0401 & 0.0466 & 0.0349 & 0.0048 & 0.0059 \\
    & RVQ(6) & 0.0349 & 0.0353 & 0.0346 & 0.0263 & 0.0302 & 0.0231 & 0.0023 & 0.0032 \\
    \midrule
    
    \multirow{5}{*}{\centering \(2^{10}\)} & VQ & 0.1529 & 0.1421 & 0.1616 & 0.0792 & 0.0928 & 0.0681 & 0.0175 & 0.0184 \\
    & FSQ & 0.1710 & 0.1498 & 0.1879 & 0.0971 & 0.1149 & 0.0828 & 0.0144 & 0.0199 \\
    & LFQ & 0.2914 & 0.2537 & 0.3218 & 0.1216 & 0.1392 & 0.1074 & 0.0273 & 0.0338 \\
    & RVQ(3) & 0.0476 & 0.0461 & 0.0488 & 0.0339 & 0.0390 & 0.0297 & 0.0030 & 0.0048 \\
    & RVQ(6) & 0.0251 & 0.0252 & 0.0249 & 0.0193 & 0.0223 & 0.0169 & 0.0017 & 0.0029 \\
    \midrule
    
    \multirow{3}{*}{\centering \(2^{11}\)} & VQ & 0.1472 & 0.1422 & 0.1513 & 0.0693 & 0.0801 & 0.0606 & 0.0188 & 0.0187 \\
    & FSQ & 0.1644 & 0.1448 & 0.1803 & 0.0972 & 0.1147 & 0.0830 & 0.0131 & 0.0198 \\
    & LFQ & 0.2689 & 0.2345 & 0.2967 & 0.1148 & 0.1316 & 0.1019 & 0.0251 & 0.0309 \\
    \midrule
    
    \multirow{3}{*}{\centering \(2^{12}\)} & VQ & 0.1375 & 0.1268 & 0.1461 & 0.0755 & 0.0879 & 0.0656 & 0.0137 & 0.0161 \\
    & FSQ & 0.1500 & 0.1427 & 0.1559 & 0.0891 & 0.1055 & 0.0758 & 0.0131 & 0.0183 \\
    & LFQ & 0.2473 & 0.2184 & 0.2692 & 0.1097 & 0.1258 & 0.0976 & 0.0229 & 0.0287 \\
    \midrule
    
    \multirow{3}{*}{\centering \(2^{14}\)} & VQ & 0.1015 & 0.0944 & 0.1073 & 0.0624 & 0.0724 & 0.0543 & 0.0084 & 0.0105 \\
    & FSQ & 0.1307 & 0.1230 & 0.1369 & 0.0831 & 0.0981 & 0.0709 & 0.0108 & 0.0136 \\
    & LFQ & 0.2215 & 0.1947 & 0.2413 & 0.0998 & 0.1146 & 0.0893 & 0.0198 & 0.0256 \\
    \midrule
    
    \multirow{3}{*}{\centering \(2^{16}\)} & VQ & 0.1071 & 0.1006 & 0.1123 & 0.0660 & 0.0770 & 0.0571 & 0.0091 & 0.0123 \\
    & FSQ & 0.1216 & 0.1194 & 0.1234 & 0.0780 & 0.0938 & 0.0652 & 0.0103 & 0.0116 \\
    & LFQ & 0.2037 & 0.1792 & 0.2236 & 0.0948 & 0.1093 & 0.0845 & 0.0179 & 0.0238 \\
    \bottomrule
    \end{tabular}
    }
    \label{tab:abl_quantizer}
\end{table*}
\begin{table*}
    \centering
    \caption{Results of \vqname with different quantizers with different codebook sizes.}
    \adjustbox{width=0.9\linewidth}{
    \begin{tabular}{ll|ccc|ccc|cc}
    
    \toprule 
     \multirow{2}{*}{\centering Size} & \multirow{2}{*}{\centering Quantizer} & \multicolumn{3}{c|}{MPJPE} & \multicolumn{3}{c|}{PA-MPJPE} & \multirow{2}{*}{\centering ADE} & \multirow{2}{*}{\centering FDE}\\
     & & all & body & hand & all & body & hand & & \\
     \midrule

     \multirow{5}{*}{\centering \(2^{9}\)} & VQ & 0.1762 & 0.1619 & 0.1877 & 0.0827 & 0.0964 & 0.0717 & 0.0182 & 0.0188 \\
    & FSQ & 0.0986 & 0.0954 & 0.1011 & 0.0608 & 0.0726 & 0.0514 & 0.0082 & 0.0119 \\
    & LFQ & 0.2721 & 0.2244 & 0.3104 & 0.1224 & 0.1413 & 0.1072 & 0.0242 & 0.0320 \\
    & RVQ(3) & 0.0611 & 0.0583 & 0.0634 & 0.0416 & 0.0481 & 0.0363 & 0.0041 & 0.0057 \\
    & RVQ(6) & 0.0319 & 0.0308 & 0.0328 & 0.0235 & 0.0268 & 0.0209 & 0.0020 & 0.0034 \\
    \midrule
    
    \multirow{5}{*}{\centering \(2^{10}\)} & VQ & 0.1523 & 0.1374 & 0.1643 & 0.0767 & 0.0894 & 0.0665 & 0.0162 & 0.0183 \\
    & FSQ & 0.0930 & 0.0887 & 0.0965 & 0.0576 & 0.0678 & 0.0493 & 0.0088 & 0.0115 \\
    & LFQ & 0.2807 & 0.2444 & 0.3102 & 0.1175 & 0.1352 & 0.1032 & 0.0297 & 0.0364 \\
    & RVQ(3) & 0.0517 & 0.0497 & 0.0533 & 0.0358 & 0.0414 & 0.0313 & 0.0037 & 0.0051 \\
    & RVQ(6) & 0.0293 & 0.0292 & 0.0294 & 0.0227 & 0.0259 & 0.0202 & 0.0019 & 0.0031 \\
    \midrule
    
    \multirow{3}{*}{\centering \(2^{11}\)} & VQ & 0.1142 & 0.1015 & 0.1245 & 0.0647 & 0.0765 & 0.0553 & 0.0125 & 0.0165 \\
    & FSQ & 0.0864 & 0.0835 & 0.0887 & 0.0546 & 0.0646 & 0.0466 & 0.0077 & 0.0105 \\
    & LFQ & 0.2684 & 0.2225 & 0.3055 & 0.1285 & 0.1455 & 0.1147 & 0.0225 & 0.0306 \\
    \midrule
    
    \multirow{3}{*}{\centering \(2^{12}\)} & VQ & 0.1027 & 0.0960 & 0.1082 & 0.0562 & 0.0661 & 0.0482 & 0.0121 & 0.0131 \\
    & FSQ & 0.0800 & 0.0765 & 0.0828 & 0.0494 & 0.0581 & 0.0423 & 0.0066 & 0.0099 \\
    & LFQ & 0.2815 & 0.2353 & 0.3187 & 0.1205 & 0.1392 & 0.1055 & 0.0312 & 0.0393 \\
    \midrule
    
    \multirow{3}{*}{\centering \(2^{14}\)} & VQ & 0.1005 & 0.0924 & 0.1070 & 0.0614 & 0.0711 & 0.0535 & 0.0092 & 0.0127 \\
    & FSQ & 0.0772 & 0.0744 & 0.0793 & 0.0489 & 0.0576 & 0.0420 & 0.0067 & 0.0090 \\
    & LFQ & 0.2538 & 0.2104 & 0.2876 & 0.1123 & 0.1298 & 0.0987 & 0.0208 & 0.0284 \\
    \midrule
    
    \multirow{3}{*}{\centering \(2^{16}\)} & VQ & 0.1187 & 0.1089 & 0.1266 & 0.0661 & 0.0779 & 0.0566 & 0.0109 & 0.0141 \\
    & FSQ & 0.0608 & 0.0592 & 0.0620 & 0.0382 & 0.0445 & 0.0330 & 0.0056 & 0.0073 \\
    & LFQ & 0.2415 & 0.1987 & 0.2743 & 0.1089 & 0.1254 & 0.0958 & 0.0192 & 0.0267 \\
    \bottomrule
    \end{tabular}
    }
    \label{tab:abl_vqpp_quantizer}
\end{table*}
\subsection{Impact of Different Quantizers and Codebook Sizes}
Quantizer is one of the core components of VQ-VAE, determining the quality and size of the VQ-VAE's codebook, which directly affects the learning difficulty and performance of the autoregressive transformer. We experimented with different codebook sizes for four types of quantizers: VQ, RVQ, FSQ, and LFQ on vanilla motion VQ. As shown in ~\tabref{tab:abl_quantizer}, we found that as the codebook size increases, FSQ consistently shows efficiency advantages but always falls short in reconstruction performance compared to VectorQuantizer. However, in \vqname, as shown in ~\tabref{tab:abl_vqpp_quantizer}, FSQ demonstrates significant advantages even at smaller codebook sizes. This indicates that the quantizer cannot function independently. When the encoder can extract high-quality latent vectors, FSQ (Finite Scalar Quantization) is a better choice; conversely, VectorQuantizer is more suitable. However, based on the experimental results, LFQ (Lattice-based Finite Quantization) is not well-suited for motion tasks. In the final version of \vqname and following experiments, we use FSQ with levels [7,5,5,5,5] to balance reconstruction performance and computational cost. 

\begin{table*}
    \centering
    \caption{Impact of network width and codebook channel dimensions on reconstruction performance.}
    \adjustbox{width=0.9\linewidth}{
    \begin{tabular}{l|ccc|ccc|cc}
    
    \toprule 
     Metrics & \multicolumn{3}{c|}{MPJPE} & \multicolumn{3}{c|}{PA-MPJPE} & \multirow{2}{*}{\centering ADE} & \multirow{2}{*}{\centering FDE}\\
    Code dim & all & body & hand & all & body & hand & & \\
    \midrule

    512 & 0.0800 & 0.0765 & 0.0828 & 0.0494 & 0.0581 & 0.0423 & 0.0066 & 0.0099 \\ 
    768 & 0.0838 & 0.0785 & 0.0880 & 0.0501 & 0.0586 & 0.0433 & 0.0085 & 0.0104 \\ 
    1024 & 0.0855 & 0.0828 & 0.0876 & 0.0512 & 0.0603 & 0.0438 & 0.0082 & 0.0104 \\ 
    1536 & 0.0846 & 0.0802 & 0.0879 & 0.0507 & 0.0594 & 0.0435 & 0.0083 & 0.0103 \\ 
    \bottomrule
    \end{tabular}
    }
    \label{tab:abl_zdim}
\end{table*} 
\subsection{Impact of dimensions}

We pay attention to the width of the network (number of channels) and the dimensionality of the codes, keeping them consistent. As shown in \tabref{tab:abl_zdim}, we find that simply increasing the dimensionality does not improve the tokenizer's performance; instead, a smaller dimensionality (512) achieves the best results.

\begin{table*}
    \centering
    \caption{Reconstruction results of motion tokenizers with different depths.}
    \adjustbox{width=0.9\linewidth}{
    \begin{tabular}{l|ccc|ccc|cc}
    
    \toprule 
     Metrics & \multicolumn{3}{c|}{MPJPE} & \multicolumn{3}{c|}{PA-MPJPE} & \multirow{2}{*}{\centering ADE} & \multirow{2}{*}{\centering FDE}\\
    Depth & all & body & hand & all & body & hand & & \\
    \midrule
    2 & 0.0902 & 0.0861 & 0.0933 & 0.0531 & 0.0626 & 0.0461 & 0.0085 & 0.0109 \\
    3 & 0.0800 & 0.0765 & 0.0828 & 0.0494 & 0.0581 & 0.0423 & 0.0066 & 0.0099 \\
    4 & 0.0753 & 0.0718 & 0.0779 & 0.0472 & 0.0556 & 0.0405 & 0.0059 & 0.0092 \\
    5 & 0.0721 & 0.0687 & 0.0746 & 0.0458 & 0.0539 & 0.0392 & 0.0054 & 0.0087 \\
    
    \bottomrule
    \end{tabular}
    }
    \label{tab:abl_depth}
\end{table*}

\section{Impact of ResNet Block Depth}
\label{subsec:depth}
We vary the number of ResNet blocks (referred to as the depth of the motion tokenizer) 
in each encoder and decoder layer. 
As shown in \tabref{tab:abl_depth}, 
the reconstruction quality of the motion tokenizer is positively correlated with the depth.

\begin{table*}
    \centering
    \caption{Comparison of the effects of different normalization and activation methods}
    \adjustbox{width=0.9\linewidth}{
    \begin{tabular}{ll|ccc|ccc|cc}
    
    \toprule 
     \multirow{2}{*}{\centering Norm} & \multirow{2}{*}{\centering Act} & \multicolumn{3}{c|}{MPJPE} & \multicolumn{3}{c|}{PA-MPJPE} & \multirow{2}{*}{\centering ADE} & \multirow{2}{*}{\centering FDE}\\
     & & all & body & hand & all & body & hand & & \\
    \midrule

    None & ReLU & 0.0800 & 0.0765 & 0.0828 & 0.0494 & 0.0581 & 0.0423 & 0.0066 & 0.0099 \\ 
    GroupNorm(32) & SiLU & 0.0796 & 0.0748 & 0.0835 & 0.0508 & 0.0596 & 0.0438 & 0.0067 & 0.0092 \\ 
    LayerNorm & SiLU & 0.0776 & 0.0741 & 0.0804 & 0.0491 & 0.0576 & 0.0421 & 0.0069 & 0.0092 \\ 
    \bottomrule
    \end{tabular}
    }
    \label{tab:abl_norm_act}
\end{table*}
 
\subsection{Impact of Different Normalization and Activation Methods}
Vanilla Motion VQ does not use normalization layers and employs ReLU as the activation function. In recent VQ-VAE-related works~\cite{vqgan,zheng2022movq,cosmos}, the more commonly used combinations are GroupNorm+SiLU or LayerNorm+SiLU. As shown in \tabref{tab:abl_norm_act}, we compare these three combinations and find that LayerNorm+SiLU performs the best, while the combination without normalization and using ReLU as the activation function performs the worst. However, we also observe that SiLU, GroupNorm, and LayerNorm significantly increase computational overhead, while the performance gains are relatively marginal. Therefore, balancing these two aspects, we ultimately choose not to use normalization and retain ReLU as the activation function.

\begin{table*}
    \centering
    \caption{Comparison between vanilla convolution and causal convolution}
    \adjustbox{width=0.9\linewidth}{
    \begin{tabular}{l|ccc|ccc|cc}
    
    \toprule 
     \multirow{2}{*}{\centering Conv} & \multicolumn{3}{c|}{MPJPE} & \multicolumn{3}{c|}{PA-MPJPE} & \multirow{2}{*}{\centering ADE} & \multirow{2}{*}{\centering FDE}\\
     & all & body & hand & all & body & hand & & \\
    \midrule

    vanilla & 0.0800 & 0.0765 & 0.0828 & 0.0494 & 0.0581 & 0.0423 & 0.0066 & 0.0099 \\ 

    causal & 0.1200 & 0.1157 & 0.1235 & 0.0700 & 0.0842 & 0.0584 & 0.0099 & 0.0132 \\ 

    \bottomrule
    \end{tabular}
    }
    \label{tab:abl_causal}
\end{table*}
\subsection{Impact of Causal Design}
In standard temporal convolutions, each frame of the output is based on both previous and future frames. 
However, in prediction tasks, this non-causal configuration can lead to inconsistencies between training and inference, 
as during inference, the convolution can only access frames up to the current time step. 
Therefore, some works~\cite{cosmos,emu3} propose using \textbf{causal convolutions} in VQ-VAE. 
We also experimented with this design in our motion tokenizer. 
As shown in \tabref{tab:abl_causal}, 
although causal convolutions are theoretically more reasonable, 
they significantly degrade the reconstruction quality of the motion tokenizer. 
Thus, we ultimately did not adopt causal convolutions.

\begin{table*}
    \centering
    \caption{Results with different patchify methods and patch sizes.}
    \adjustbox{width=0.9\linewidth}{
    \begin{tabular}{ll|ccc|ccc|cc}
    
    \toprule 
    \multirow{2}{*}{\centering Method} & \multirow{2}{*}{\centering PS} & \multicolumn{3}{c|}{MPJPE} & \multicolumn{3}{c|}{PA-MPJPE} & \multirow{2}{*}{\centering ADE} & \multirow{2}{*}{\centering FDE}\\
     & & all & body & hand & all & body & hand & & \\
    \midrule
    haar & 1 & 0.0821 & 0.0790 & 0.0847 & 0.0500 & 0.0590 & 0.0428 & 0.0081 & 0.0109 \\
    haar & 2 & 0.0843 & 0.0805 & 0.0873 & 0.0531 & 0.0626 & 0.0453 & 0.0070 & 0.0103 \\
    haar & 4 & 0.0875 & 0.0832 & 0.0908 & 0.0552 & 0.0651 & 0.0471 & 0.0093 & 0.0115 \\
    rear & 1 & 0.0800 & 0.0765 & 0.0828 & 0.0494 & 0.0581 & 0.0423 & 0.0066 & 0.0099 \\
    rear & 2 & 0.0818 & 0.0779 & 0.0849 & 0.0513 & 0.0607 & 0.0437 & 0.0070 & 0.0099 \\
    rear & 4 & 0.0868 & 0.0847 & 0.0885 & 0.0531 & 0.0627 & 0.0455 & 0.0107 & 0.0118 \\
    \bottomrule
    \end{tabular}
    }
    \label{tab:abl_patchify}
\end{table*}

\subsection{Impact of Patchify Methods}
Apart from downsampling layers, another common method to reduce the temporal length of input sequences is patchify: reducing the time dimension by transferring it to the channel dimension at the input stage. Cosmos proposes two different patch methods. The first method stacks multi-frame signals along the channel dimension (implemented as a simple rearrange operation, denoted as rear), thereby reducing the temporal sequence length. The second method applies a Haar wavelet transform at the input stage (denoted as haar). We compare the results of different patch methods and patch sizes with a temporal compression rate of 4 in ~\ref{tab:abl_patchify}. We find that increasing the patch size effectively reduces computational overhead (since ResNet blocks operate on shorter temporal sequences) but sacrifices some reconstruction accuracy. Additionally, the Haar wavelet transform is not more effective than simple multi-frame stacking in the motion domain.

\begin{table*}
    \centering
    \caption{Comparison of reconstruction quality when training with motion clips of varying lengths.}
    \adjustbox{width=0.9\linewidth}{
    \begin{tabular}{l|ccc|ccc|cc}
    
    \toprule 
     \multirow{2}{*}{\centering Length} & \multicolumn{3}{c|}{MPJPE} & \multicolumn{3}{c|}{PA-MPJPE} & \multirow{2}{*}{\centering ADE} & \multirow{2}{*}{\centering FDE}\\
     & all & body & hand & all & body & hand & & \\
    \midrule

    64 & 0.0800 & 0.0765 & 0.0828 & 0.0494 & 0.0581 & 0.0423 & 0.0066 & 0.0099 \\

    128 & 0.0689 & 0.0658 & 0.0715 & 0.0439 & 0.0517 & 0.0377 & 0.0056 & 0.0082 \\
    
    192 & 0.0684 & 0.0640 & 0.0720 & 0.0438 & 0.0511 & 0.0378 & 0.0061 & 0.0089 \\
   
    256 & 0.0655 & 0.0629 & 0.0676 & 0.0417 & 0.0494 & 0.0355 & 0.0056 & 0.0081 \\

    320 & 0.0679 & 0.0643 & 0.0708 & 0.0431 & 0.0505 & 0.0371 & 0.0060 & 0.0081 \\
    
    full & 0.0681 & 0.0643 & 0.0712 & 0.0431 & 0.0503 & 0.0373 & 0.0068 & 0.0093 \\
    \bottomrule
    \end{tabular}
    }
    \label{tab:abl_length}
\end{table*}
 
\subsection{Impact of Training Segment Length}
Previous works have uniformly segmented motion sequences into fixed-length clips of 64 frames for training purposes. In our practice, however, we discovered that a VQ-VAE trained on such short clips underperforms when applied to complete motion sequences. Consequently, we experimented with clips of varying frame lengths, as well as utilizing the entire motion sequence in its unsegmented form (denoted as 'full'). As shown in ~\tabref{tab:abl_length}, we find that setting the maximum length of the motion sequence to \textbf{256 frames} 
yields the best reconstruction performance for the motion tokenizer.

\section{Evaluation Metrics}

\label{sec:add_eval_metrics}

In this section, we elucidate the computational methodologies for the evaluation metrics associated with each task. 

\noindent \textbf{Text-to-Motion.} Following ~\cite{t2m}, We assess the quality of generated motions using Fréchet Inception Distance (FID), evaluate the alignment between motions and textual descriptions with MultiModal Distance and R-Precision, and measure the diversity of the motions using Diversity metrics. FID quantifies the distance between the statistical distributions of the generated motions and the ground truth motions. Multimodal Distance evaluates the distance between the feature vectors of the generated motions and their corresponding conditional captions. Conversely, R-Precision measures the proportion of generated motions within a batch of size \(B\) whose feature vectors rank among the top \(k\) closest to those of their respective conditional captions, the calculation process can be expressed as follow:
\begin{equation}
\text{R-Precision} = \frac{1}{B} \sum_{i=1}^{B} \mathbb{I}\left( \mathbf{m}_i \in \text{Top}_k\left(\{\mathbf{m}_j\}_{j=1}^{B}, \mathbf{c}_i\right) \right)
\end{equation}

Based on the aforementioned descriptions, the computation of these metrics relies on motion latents that effectively capture motion quality and align with textual descriptions. For single-person motion and text data, we follow ~\cite{humantomato,petrovich2023tmr}, employing contrastive learning techniques to align motion latents with text latents. In the case of two-person motions, we concatenate the two motions along the channel dimension to serve as the motion input to TMR, while using dual-person motion descriptions as the textual input. We trained the TMR model on the MotionHub dataset using single-person motions for 1,300 epochs and dual-person motions for 1,500 epochs, respectively.

The Diversity metric quantifies the variability among the actions generated by the model. To compute this metric, we randomly select 300 motion samples from all generated motions, extract their feature vectors, and calculate the average L2 distance between these samples.

\noindent \textbf{Motion-to-Text.} We employ R-Precision and Multimodal Distance to evaluate the alignment between ground truth motions and model-generated captions, utilizing the same computation methods as those applied in text-to-motion tasks. To assess the similarity between the generated captions and the ground truth captions, we utilize traditional NLP metrics, including BLEU~\cite{bleu}, ROUGE-L~\cite{rouge}, CIDEr-D~\cite{cider}, and BERTScore~\cite{bertscore}.

\noindent \textbf{Music-to-Dance.} For dance motions, we adhere to previous works by utilizing FairMotion~\cite{fairmotion} to extract the kinetic features of the motions. The Kinetic Fréchet Inception Distance (Kinetic FID, \(FID_k\) measures the statistical distance between the kinetic features of generated dance motions and those of ground truth motions to evaluate dance quality. \(Div_k\) computes the pairwise L2 distances between the kinetic features of all generated dance motions to assess their diversity. The Beat Alignment Score (BAS) calculates the average L2 error between the musical beats and the motion beats, as defined by the following formula:
\begin{equation}
BAS = \frac{1}{|B_m|} \sum_{t_m \in B_m} \exp \left\{ -\frac{\min_{t_d \in B_d} \|t^d - t^m\|^2}{2\sigma^2} \right\}
\end{equation}
In this framework, \(t_d\) denotes the beats of the dance motions, calculated as the maximum local velocity of the dance movements. \(t_m\) represents the beats of the music, which are extracted using the Librosa~\cite{librosa} library. However, the BAS metric has certain limitations, as it does not penalize dance motions with an excessive number of beats. Consequently, in some cases, the generated dance motions may exhibit higher BAS scores than the ground truth.

\noindent \textbf{Dance-to-Music.} Following ~\cite{loris,m3gpt}, we assess the beat accuracy of the generated music. To evaluate the alignment between the kinematic beats of the generated dances and the input music beats, we employ two metrics. Let \( B_m \) denote the total number of musical beats, \( B_k \) represent the total number of kinematic beats, and \( B_a \) indicate the number of kinematic beats that are aligned with musical beats. According to the implementation of LORIS~\cite{loris}, multiple beats occurring within a single second are treated as one. The two metrics are defined as follows:
\begin{itemize}
    \item \textbf{Beat Coverage} (\( B_k / B_m \)): Measures the proportion of kinematic beats relative to musical beats.
    \item \textbf{Beat Hit Rate} (\( B_a / B_k \)): Quantifies the ratio of aligned kinematic beats to total kinematic beats.
\end{itemize} 

Consistent with ~\cite{loris,m3gpt}, we compute the F1 score for Beat Coverage Score (BCS) and Beat Hit Score (BHS) using the following formula:
\begin{equation}
F1 = \frac{2 \cdot \text{BCS} \cdot \text{BHS}}{\text{BCS} + \text{BHS}}
\end{equation}

\noindent \textbf{Speech-to-Gesture.} Following ~\cite{emage}, we use FGD to evaluate the quality of gesture motions, L1Div to measure the diversity of gesture motions, and Beat Alignment to assess the rhythm consistency between gesture motions and speech. The computation of L1Div is as follows: L1Div measures the diversity of gesture motions by calculating the average absolute deviation of all results from their mean. To extract gesture latents for computing FGD, we trained a Gesture VAE~\cite{temos} on the BEATv2~\cite{emage} dataset.

\section{Visual Demonstration}
\label{sec:demo}

To enhance interpretability and provide comprehensive insights, we present visualizations of benchmark examples from MotionHub in ~\figref{fig:motionhub}. These visualizations serve as a foundation for understanding the capabilities and outputs of our framework.

Furthermore, we showcase detailed demonstrations of key motion synthesis tasks, including:
\begin{itemize}
    \item Single-agent Text-to-Motion ~\figref{fig:demo_t2m}
    \item Dual-agent Text-to-Motion ~\figref{fig:demo_t2m_2p}
    \item Music-to-Dance ~\figref{fig:demo_m2d}
    \item Speech-to-Gesture ~\figref{fig:demo_s2g}
    \item Motion-Prediction ~\figref{fig:demo_pred}
    \item Motion-Inbetween ~\figref{fig:demo_inbetween}
\end{itemize}

These visualizations highlight the versatility and performance of our approach across diverse motion-related tasks.

\section{Discussion and Future Work}

While \name achieves strong performance across diverse motion tasks, several limitations suggest avenues for further research:

\textbf{Data Scale and Diversity.}  
Despite the size of MotionHub, motion data remains scarce compared to text, image, and video domains.  High-precision motion capture is costly and often confined to single‐agent or dyadic interactions.  Future efforts will expand MotionHub with additional multi-agent scenarios and richer audio–motion pairings—particularly crowd scenes, group choreography, and conversational gestures—to better match model capacity and scaling requirements.

\textbf{Leveraging Caption Granularity.}  
We categorize captions into macro, meso, and micro levels, but our current models do not explicitly exploit this hierarchy.  Developing methods to integrate granularity information—such as curriculum learning from macro to micro or multi‐scale attention mechanisms—could substantially improve semantic grounding and fine‐grained motion understanding.

\textbf{Extending Multi‑Agent Validation.}  
Although \name’s architecture can in principle handle any number of agents, empirical evaluation has been limited to two‐person interactions due to data constraints.  We plan to curate and incorporate larger multi-agent datasets—spanning sports teams, ensemble dance, and social gatherings—and to benchmark \name on scenarios with varying group sizes and interaction complexities.

By addressing these challenges, future versions of \name can achieve greater robustness, fidelity, and applicability in real-world, large‐scale multimodal motion understanding and generation.

\end{document}